\documentclass[runningheads]{llncs}
\usepackage{graphicx}
\usepackage{amsmath,amssymb} 
\usepackage{color}
\usepackage[width=122mm,left=12mm,paperwidth=146mm,height=193mm,top=12mm,paperheight=217mm]{geometry}
\begin{document}
\mainmatter
\def\ECCV18SubNumber{230}  

\title{HyperFusion-Net: Densely Reflective Fusion for Salient Object Detection} 
\titlerunning{ECCV-18 submission HyperFusion-Net}

\authorrunning{ECCV-18 submission HyperFusion-Net}

\author{Pingping Zhang$^{\dagger\ddagger}$\quad Huchuan Lu$^{\dagger}$\thanks{Prof. Lu is the corresponding author.}\quad Chunhua Shen$^{\ddagger}$\\
}
\institute{$^{\dagger}$Dalian University of Technology\quad\quad $^{\ddagger}$ The University of Adelaide}
\maketitle

\begin{abstract}
%
Salient object detection (SOD), which aims to find the most important region of interest and segment the relevant object/item in that area, is an important yet challenging vision task.
This problem is inspired by the fact that human seems to perceive main scene elements with high priorities.
Thus, accurate detection of salient objects in complex scenes is critical for human-computer interaction.
In this paper, we present a novel feature learning framework for SOD, in which we cast the SOD as a pixel-wise classification problem.
The proposed framework utilizes a densely hierarchical feature fusion network, named \emph{HyperFusion-Net}, automatically predicts the most important area and segments the associated objects in an end-to-end manner.
Specifically, inspired by the human perception system and image reflection separation, we first decompose input images into reflective image pairs by content-preserving transforms.
Then, the complementary information of reflective image pairs is jointly extracted by an interweaved convolutional neural network (ICNN) and hierarchically combined with a hyper-dense fusion mechanism.
Based on the fused multi-scale features, our method finally achieves a promising way of predicting SOD.
As shown in our extensive experiments, the proposed method consistently outperforms other state-of-the-art methods on seven public datasets with a large margin.
\end{abstract}

\section{Introduction}
\vspace{-2mm}
Salient object detection (SOD) aims to detect and segment the attractive objects to human observers in an image, without any prior
knowledge of image content.
It is widely used as a fundamental and useful pre-processing method for numerous object-related applications, including image compression~\cite{hadizadeh2014saliency}, information retrieval~\cite{he2012mobile,gao20123}, semantic segmentation~\cite{donoser2009saliency} and photo editing~\cite{chen2015improved}.

In the past decades, a large amount of SOD methods have been proposed~\cite{borji2015salient}.
Most of these methods adopt handcrafted visual features in detection.
Color feature is explored in various means, such as color contrast and correlation, because human vision
system is highly sensitive to color information~\cite{cheng2015global}.
Location cue, especially center-bias, is also frequently used to improve saliency detection performance, for people prefer to locate the salient objects near the center position when taking a photo~\cite{ren2014important}.
\begin{figure*}
\begin{center}
\begin{tabular}{@{}c@{}c@{}c@{}c@{}c@{}c@{}c@{}c}
\includegraphics[width=0.16\linewidth,height=1.6cm]{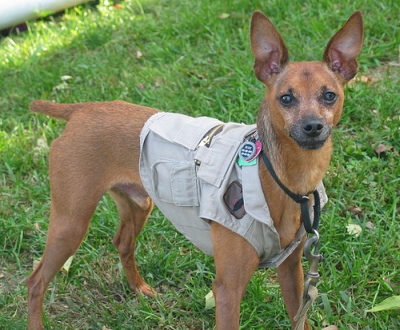} \ &
\includegraphics[width=0.16\linewidth,height=1.6cm]{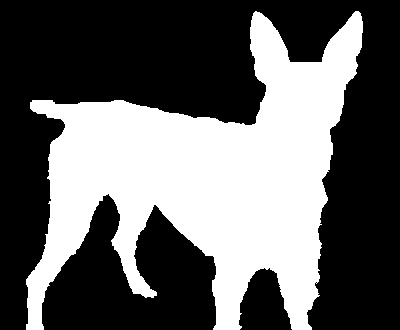} \ &
\includegraphics[width=0.16\linewidth,height=1.6cm]{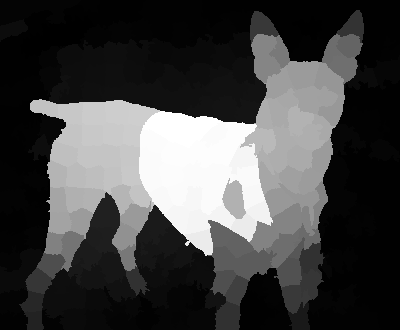} \ &
\includegraphics[width=0.16\linewidth,height=1.6cm]{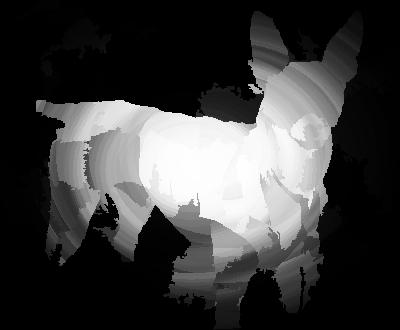} \ &
\includegraphics[width=0.16\linewidth,height=1.6cm]{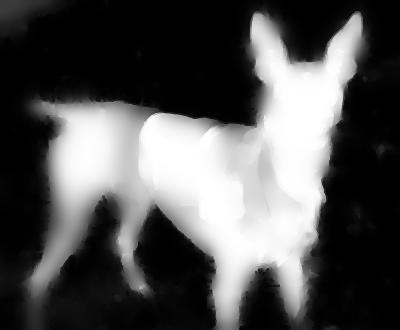} \ &
\includegraphics[width=0.16\linewidth,height=1.6cm]{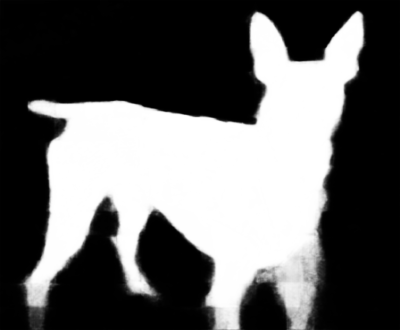} \ \\
\end{tabular}
\vspace{-4mm}
\caption{Examples of pixel-wise saliency prediction. (a) Input image. (b) Ground-truth. (c) Color cue~\cite{qin2015saliency}. (d) Location cue~\cite{li2013saliency}. (e) Deep feature~\cite{wang2016saliency}. (f) Hyper-Fusion feature.
\label{fig:cues}}
\vspace{-14mm}
\end{center}
\end{figure*}
Recently, with the advances of deep learning, learned features as saliency cues are frequently used for SOD, since
learned features have strong ability to successfully avoid the drawbacks of handcrafted features.
However, using a single cue only provides partial information of salient objects, which may lead to inaccurate detection results.
As shown in Fig.~\ref{fig:cues} (c)-(e), one can find that the saliency maps generated by a single cue may omit some salient regions
(Fig.~\ref{fig:cues} (c)-(d)) or bring in insignificant regions (Fig.~\ref{fig:cues} (e)).
Hence, it is reasonable to combine multiple cues to improve SOD results (Fig.~\ref{fig:cues} (f)).

Recent works~\cite{wang2016saliency,li2016ds,liu2016dhsnet,zhang2017amulet,hou2017deeply} also show that SOD with multi-scale features generally achieves better performance than that with single-scale one.
Different features can represent different characteristics of salient objects, and utilizing different features effectively will have positive effects on SOD.
Meanwhile, the advances of deep convolutional neural networks (CNN) enable researchers to develop various SOD methods to cooperate with multiple features.
However, even if these methods achieve very encouraging performances, there still exist some intrinsic problems.
Firstly, these methods directly encode the multi-scale features over the original input images, by which way human perception information is ignored and the SOD performance can be compromised.
Moreover, when the training data increases in number, the jointly-encoding process can be very time-consuming.
Thirdly, these methods ignore some semantic relationships among the features, which can boost the SOD performance.
Thus, coarsely utilizing all the features not only adds extra computation burden, but also prevents further improvement.

\begin{figure*}
\begin{center}
\vspace{-8mm}
\includegraphics[width=0.98\linewidth,height=6.2cm]{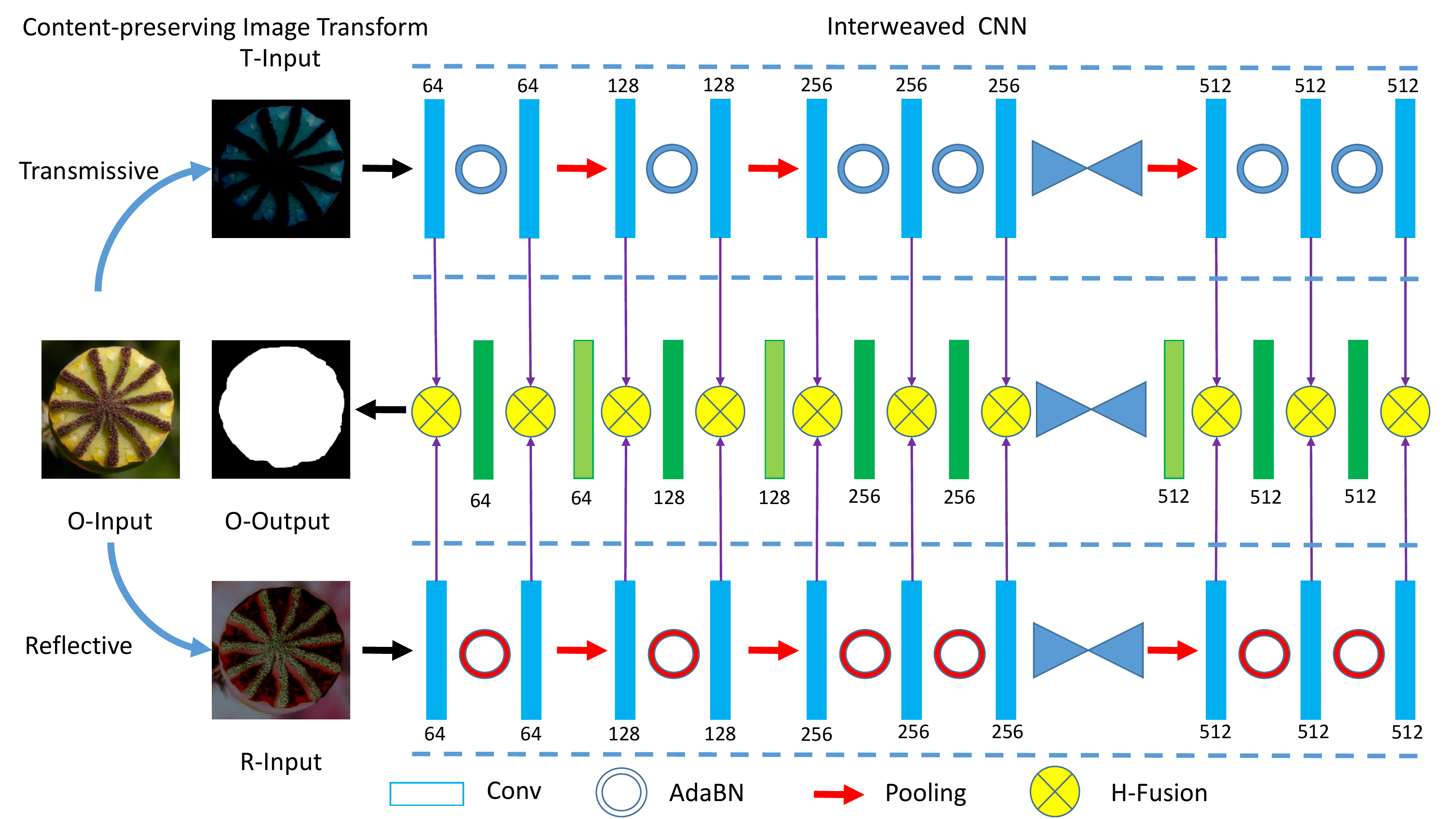}
\end{center}
\vspace{-6mm}
\caption{An overview of our SOD approach based on the VGG-16 model~\cite{simonyan2014very}.
Bottom: The weight-stitching branch for the reflected image.
Top: The weight-stitching branch for the transmitted image.
Middle: The hyper-fusion branch to densely fuse the multi-level features.
More details can be found in the main text.}
\label{fig:framework}
\vspace{-12mm}
\end{figure*}

To address above issues, we cast SOD as a pixel-wise classification task, and propose to solve complementary feature extraction and saliency region classification within a unified framework, as illustrated in Fig.~\ref{fig:framework}.
We fuse the multi-scale features into a more preferable presentation, which is more compact and discriminative for better SOD performance.
Specifically, inspired by the human perception system and image reflection separation, we first decompose input images into reflective image pairs by content-preserving transforms.
Then, we design an interweaved CNN (ICNN) which consists of two weight-stitching branches and one hyper-fusion branch.
The complementary features of reflective image pairs are jointly extracted by the proposed ICNN and hierarchically combined with a hyper-dense fusion mechanism.
Based on the fused multi-scale features, our method finally achieves a promising way of predicting SOD in an end-to-end manner.
In this manner, our proposed model sufficiently captures the clear boundaries and spatial contexts of salient objects, hence significantly boosts the performance of SOD.
We evaluate our model by comparing it with other state-of-the-art approaches on seven public benchmarks, and the experimental results demonstrate the effectiveness of our approach.

In summary, \textbf{our contributions} are three folds:
\begin{itemize}
\item
We present a novel network architecture, \emph{i.e., HyperFusion-Net}, which is specifically designed to learn complementary visual features in a fusing view and predict accurate saliency maps with human perception mechanism.
\item
We propose a hyper-dense fusion method to diversify the contributions of multi-scale features from global and local perspectives.
This fusion method is able to to learn clear object boundaries and spatially consistent saliency.
%
\item
Extensive experiments on seven large-scale saliency benchmarks demonstrate that the proposed approach achieves superior performance and outperforms the very recent state-of-the-art methods by a large margin.
\end{itemize}
\section{Related Work}
\vspace{-2mm}
\subsection{Salient Object Detection}
\vspace{-2mm}
Over the past two decades, a large mount of SOD methods have been developed.
The majority of existing methods are based on hand-crafted features.
A complete survey of these methods is beyond the scope of this paper and we refer the readers to a recent survey paper~\cite{borji2015salient} for details.
Here, we mainly focus on discussing recent methods based on deep learning architectures.

Recent years, deep learning based methods have achieved solid performance improvements in SOD.
For example, Wang \emph{et al}~\cite{wang2015deep} integrate both local pixel estimation and global proposal search for SOD by training two deep neural networks.
Zhao \emph{et al}~\cite{zhao2015saliency} propose a multi-context deep CNN framework to benefit from the local context and global context of salient objects.
Li \emph{et al}~\cite{li2015visual} employ multiple deep CNNs to extract multi-scale features for saliency prediction.
Then they propose a deep contrast network to combine a pixel-level stream and segment-wise stream for saliency estimation~\cite{li2016dcl}.
Inspired by the great success of fully convolutional networks (FCNs)~\cite{long2015fully}, Wang \emph{et al}~\cite{wang2016saliency} develop a recurrent FCN to incorporate saliency priors for more accurate saliency map inference.
Liu \emph{et al}~\cite{liu2016dhsnet} also design a deep hierarchical network to learn a coarse global estimation and then refine the saliency map hierarchically and progressively.
Then, Hou \emph{et al}~\cite{hou2017deeply} introduce dense short connections to the skip-layers within the holistically-nested edge detection (HED) architecture~\cite{xie2015holistically} to get rich multi-scale features for SOD.
Zhang \emph{et al}~\cite{zhang2017amulet} propose a bidirectional learning framework to aggregate multi-level convolutional features for SOD.
And they also develop a novel dropout to learn the deep uncertain convolutional features to enhance the robustness and accuracy of saliency detection~\cite{zhang2017learning}.
Wang \emph{et al}~\cite{wang2017stagewise} provide a stage-wise refinement framework to gradually get accurate saliency detection results.
Despite these approaches employ powerful CNNs and make remarkable success in SOD, there still exist some obvious problems.
For example, most existing methods are based on the direct supervised learning and ignore human perception mechanism.
And the fusing strategies of multiple features are sparse and insufficient.
As a result, there is still a large space for performance improvements.
We argue that a dense fusion framework with diversified fusion points and more adaptive fusion paths is in demand, which not only facilitates the gradient-based optimization process, but also provides a platform for incorporating a multi-scale understanding into the fusion process.
\vspace{-4mm}
\begin{figure*}
\begin{center}
\includegraphics[width=0.96\linewidth,height=6.4cm]{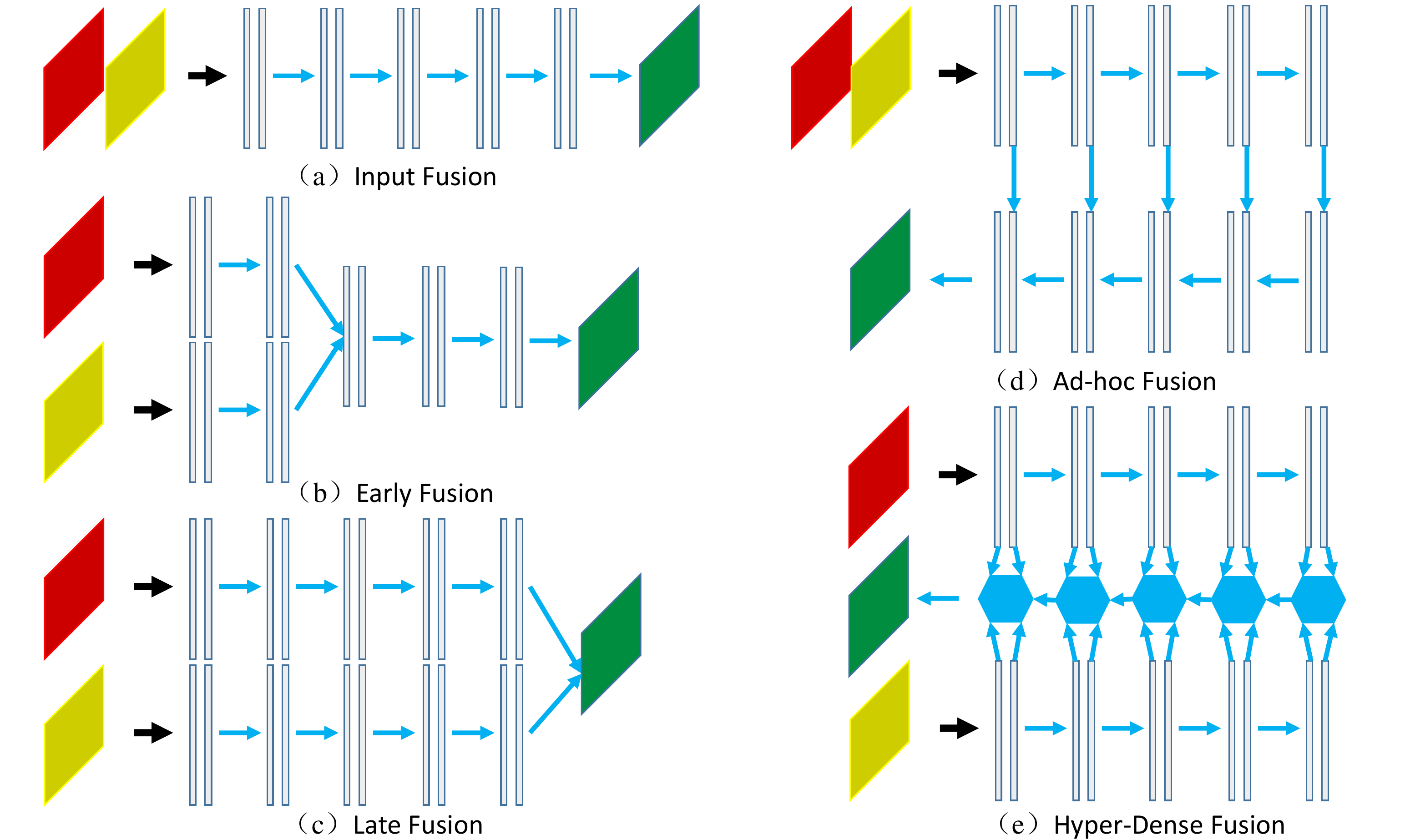}
\end{center}
\vspace{-6mm}
\caption{Different network structures for deep feature fusion.}
\label{fig:fusion}
\end{figure*}
\vspace{-12mm}
\subsection{Deep Feature Fusion}
Recently, deep CNNs have been successfully applied to various computer vision tasks due to its power in exploring multi-level representations.
Encouraged by its strengths, researchers~\cite{wang2016saliency,li2016ds,liu2016dhsnet,wang2015deep,zhao2015saliency,li2015visual,li2016dcl,lee2016deep} start to leverage CNNs to fuse multi-level or multi-cue features automatically for performance improvement.
The information from different sources is typically combined with the input fusion or early fusion or late fusion stage (shown in Fig.~\ref{fig:fusion} (a), (b), (c), respectively) via a single fusion point.
Some more ad-hoc fusion methods~\cite{zhang2017amulet,hou2017deeply} (shown in Fig.~\ref{fig:fusion} (d)) have also been introduced by considering the relationship between different scales and levels.
Unfortunately, they do not go beyond the traditional philosophy for feature fusion, which means applying existing standard methods to multiple features separately and then fusing their results in the decision stage.
To sum up, though encouraging results have been achieved, the fusion methods in previous models are typically focalized
in sparse points, which may be deficient to merge all the useful information from multiple sources.
As a result, the fusion process is brute-force and insufficient.
Different from previous methods, we argue that dense points (shown in Fig.~\ref{fig:fusion} (e)) can be applied to the feature fusion problem to enrich the fusion process, while few works take this fact into account.
Moreover, we observe that human vision system comprehends a scene in a coarse-to-fine way~\cite{allman1985stimulus}, which includes coarse understanding for identifying the location and shape of the target object, and fine capturing for exploring its detailed parts.
Similarly, the feature fusion also needs the collaborations of coarse and fine perspectives.
Thus, in this paper we fuse the multi-scale features in the coarse-to-fine manner.
\vspace{-4mm}
\section{Proposed Model}
\vspace{-2mm}
Fig.~\ref{fig:framework} illustrates the overall flowchart of our SOD method.
Inspired by human vision system, we first convert an input RGB image into a reflective image pair, \emph{i.e.}, the transmitted image (T-Input) and the reflected image (R-Input), by utilizing content-preserving transforms.
Then the image pair is fed into the weight-stitching branches of our proposed ICNN, extracting multi-level deep features.
Afterwards, the hyper-fusion branch hierarchically integrates the complementary features into the same resolution of input images.
Finally, the saliency map is predicted by exploiting integrated features.
In the following subsections, we will elaborate the proposed image separation, ICNN architecture and the hyper-fusion method in detail.
\vspace{-4mm}
\subsection{Content-preserving Image Separation}
Essentially, human vision system understands environments from 3D perception.
Image separation plays an important role in the perception process~\cite{hanson1978computer,levin2007user,li2014single}.
When image scenes are separated adequately, existing computer vision algorithms can better understand image contents since other irrelevant backgrounds are decreased.
Motivated by this fact, we resolve the SOD problem in the simple human perception and image separation views.
To be specific, we pose the image separation as a content-preserving image transformation task, for which we transform an input image into different visual domains.
We first convert the original RGB image $X_{O}\in R^{W\times H\times 3}$ to a reflective image pair by the following specular reflection function,
\begin{align}
Sep(X_{O},k) &= (X_{O}-E,\phi(X_{O}-E,k))),\\
&= (X_{O}-E,-k(X_{O}-E))\\
&= (X_T, X^{k}_R).
  \label{equ:equ1}
\end{align}
\vspace{-2mm}
where $\phi$ is a content-preserving transformer, $k$ is a hyperparameter to control the reflection scale and $E\in R^{W\times H\times 3}$ is the mean of an image or image dataset.
%
To reduce the computation, in this paper we use $k=1$ and the mean of the ImageNet dataset~\cite{imagenet_cvpr09}.
From above equations, one can see that the converted image pair, \emph{i.e.}, $X_T$ and $X^{k}_R$, is reciprocal with a reflection plane.
In detail, the reflection scheme is a pixel-wise negation operator, allowing the given images to be reflected in both positive and negative directions while maintaining the same content of images, as shown in Fig.~\ref{fig:framework}.
In the proposed reflection, we use the scale operator to implement the reflection, however, it is not the only feasible method.
For example, the reflection can be other non-linear operators, such as quadratic form, exponential transform and
logarithmic transform, to add more diversity.
By transforming images, the proposed algorithm makes a key difference from previous SOD methods as plausible reflected scenes can be obtained, which are based on the optical aberration and human perception.
%
%
%
In addition, different from previous image separation methods~\cite{levin2007user,li2014single,shen2008chromaticity}, our method does not rely on a certain approximation model of the reflection as it may restrict the algorithm to a specific case.
Instead, we leverage the fact that an observed image contains contents of the transmitted scene and reflected scene.
It leads us to model an observed image using a feature space instead of a pixel-level combination.
The network can also be trained in a multi-source manner by taking transmitted and reflected images as input.
\vspace{-2mm}
\subsection{Joint Feature Extraction by ICNN}
In order to extract the complementary information from the separated views, we propose an interweaved CNN which consists of two weight-stitching branches to extract multi-level features and one hyper-fusion branch to combine them.
%
%
%
More specifically, we build the two weight-stitching branch, following the VGG-16 model~\cite{simonyan2014very}.
Each weight-stitching branch has 13 convolutional layers (kernel size = $3\times 3$, stride size = 1) and 4 max pooling layers (pooling size = $2\times 2$, stride = 2).
For notational simplicity, we refer to the ConvNet as a function $f_{CNN}(X; \theta)$, that takes $X$ as input and $\theta$ as parameters.
The ICNN output multi-level feature maps with different sizes as the representations of the input image pair generated from above content-preserving transforms.
We denote the joint feature extraction process as follows:
\begin{align}
\{f^{l}_{T},f^{l}_{R}\} &= \{f^{l}_{ICNN}(X_{T}; \theta^{conv}_{ws}, \theta^{bn}_{T}),f^{l}_{ICNN}(X_{R}; \theta^{conv}_{ws}, \theta^{bn}_{R})\},
\label{equ:equ2}
\end{align}
where $f^{l}_{T}$ and $f^{l}_{R}$ denote the $l$-layer feature representation of images
$X_{T}$ and $X_{R}$, respectively.
\{$\cdot$,$\cdot$\} is the concatenation operator in channel-wise.
$\theta^{conv}_{ws}$ are the shared parameters of the convolutional layers in the two weight-stitching branches.
Note that, the weight-stitching branches are designed to share weights in convolutional layers, but with the adaptive batch normalization (AdaBN)~\cite{li2016revisiting}.
In other words, we keep the weights of corresponding convolutional layers of the two weight-stitching branches the same, while use different learnable BN (\emph{i.e.}, $\theta^{bn}_{T}$ and $\theta^{bn}_{T}$) between the convolution and ReLU operators~\cite{zhang2017amulet}.
The main reason of this design is that after performing the content-preserving transform, the reflective images have different image domains.
Domain related knowledge heavily affects the statistics of BN layers.
In order to learn domain invariant features, it is beneficial for each domain to keep its own BN statistics in each layers.
Through the two weight-stitching branches, our model learns two complementary groups of features that we successively leverage for the hyper-dense feature fusion.
%
%
In addition, according to the philosophy introduced in~\cite{Hou2017DualNet}, the proposed architecture learns two sets of complementary features more discriminative thanks to the different transmitted and reflected modalities.
%
%
%
%
%
%
\vspace{-2mm}
\subsection{Hyper-Densely Hierarchical Fusion}
Inspired by the recent success of DenseNets~\cite{huang2017densely}, we leverage a novel hyper-densely connected pattern to address the feature fusion problem.
To be specific, we propose a hyper-dense architecture, named H-Fusion (Hyper-Densely Hierarchical Fusion), for the multi-level features of image pairs, as shown in Fig.~\ref{fig:fusion} (e).
In DenseNets, connectivity in each block follows a pattern that iteratively concatenates all feature outputs in a feed-forward manner, \emph{i.e.},
\begin{align}
f^{l} &= g(\{f^{1}_{CNN}(X; \theta), f^{2}_{CNN}(X; \theta),..., f^{l-2}_{CNN}(X; \theta),f^{l-1}_{CNN}(X; \theta)\}),
\label{equ:equ3}
\end{align}
where $g$ is the fusion function, typically a convolution followed by a non-linear activation function.
Unlike DenseNets, where dense connections are employed through all the layers in a single stream, we exploit the concept of dense connectivity in a multi-source image setting.
Each information source is integrated through its dedicated module and the extracted descriptors are then concatenated to perform the final classification.
In this scenario, dense connections occur not only between layers within the same path, but also between layers in different paths.
Formally, the hyper-dense fusion architecture is defined by
\begin{equation}
  \label{equ:equ4}
\hat{f}^{l}=
\left\{
\begin{aligned}
&g(\{f^{l}_{T}, \hat{f}^{l+1}, f^{l}_{R}\}; \theta_{hf}),~\underline{L}_{m} \le l < \overline{L}_{m}\\
&g(\{f^{l}_{T}, f^{l}_{R}\}; \theta_{hf}),~l = \overline{L}_{m}
\end{aligned}
\right.
\end{equation}
\begin{align}
\tilde{f}^{l} &= h(\{\hat{f}^{\underline{L}_{1}}, \hat{f}^{\underline{L}_{2}},..., \hat{f}^{\underline{L}_{m-1}}, \hat{f}^{\underline{L}_{m}}; \theta_{hf}\}), m \in M,
\label{equ:equ5}
\end{align}
where $\hat{f}^{l}$ and $\tilde{f}^{l}$ are integrated features at the $l$-th layer with the same and different resolution, respectively.
$\theta_{hf}$ is the parameter of the hyper-fusion branch.
$\underline{L}_{m}$ and $\overline{L}_{m}$ are the layer bound of the $m$-th block.
$h$ denotes the integration operator, which is a $1\times 1$ convolutional layer followed by a deconvolutional layer, to ensure the same resolution.
Setting up the fusion process in Equ.~\ref{equ:equ4}, which takes both data sources into account at the same time, ensures that we can merge complementary and useful features for SOD.
In other words, our H-Fusion considers a more sophisticated connectivity pattern that also links the output from layers in different streams, each one associated with a different image modality.
In addition, to preserve the spatial structure and enhance the contextual information, we integrate the multi-level reflection features in a hierarchical manner, quite different from the DenseNets.

Based on the fused features, we adhere an additional convolutional layer to the ICNN for the saliency map prediction.
The numbers in Fig.~\ref{fig:framework} illustrate the detailed filter setting in each convolutional layer.
\subsection{Network Training and Testing}
Given a training dataset $ S=\{(X_n,Y_n)\}^{N}_{n=1}$ with $N$ training pairs, where $X_n =
\{x^n_i,i = 1,...,T\}$ and $Y_n = \{y^n_i,i = 1,...,T\}$ are the input image and the binary ground-truth image with $T$ pixels, respectively.
$y^n_i = 1$ denotes the foreground pixel and $y^n_i = 0$ denotes the background pixel.
For notional simplicity, we subsequently drop the subscript $n$ and consider each image independently.

In most of existing deep learning based SOD methods, the loss function used to train the network is the standard pixel-wise binary cross-entropy (BCE) loss:
\begin{equation}
  \label{equ:equ6}
\begin{aligned}
  \mathcal{L}_{bce}= -\sum_{i\in Y_{+}} \text{log~Pr}(y_{i}=1|X;\theta)-\sum_{i\in Y_{-}} \text{log~Pr}(y_{i}=0|X;\theta).
\end{aligned}
\end{equation}
where $\theta$ is the parameter of the overall network.
Pr$(y_i =1|X;\theta)\in [0,1]$ is the confidence score of the network prediction that measures how likely the pixel belong to the foreground.
$Y_{+}$ and $Y_{-}$ denote the foreground and background pixel sets in ground truth, respectively.

However, for a typical natural image, the class distribution of salient/non-salient pixels is heavily imbalanced: most of the pixels in the ground truth are non-salient.
To automatically balance the loss between positive/negative classes, we introduce a class-balancing weight $\beta$ on a per-pixel term basis, following~\cite{xie2015holistically}.
Specifically, we define the following weighted cross-entropy loss function,
\begin{equation}
  \label{equ:equ7}
\begin{aligned}
  \mathcal{L}_{wbce}= - \beta \sum_{i\in Y_{+}} \text{log~Pr}(y_{i}=1|X;\theta)-(1-\beta)\sum_{i\in Y_{-}} \text{log~Pr}(y_{i}=0|X;\theta).
\end{aligned}
\end{equation}
The loss weight $\beta = |Y_{-}|/|Y|$, and $|Y_{+}|$ and $|Y_{-}|$ denote the foreground and background pixel number, respectively.
In addItion, it is also crucial to preserve the overall spatial structure of salient objects.
Thus, we also minimize the structure perceptual (SP) loss~\cite{johnson2016perceptual},
\begin{align}
\mathcal{L}_{sp} = \sum_{l=1}^{L}\lambda_{l}||\phi_{l}(Y;w)-\phi_{l}(\hat{Y};w)||_2,
  \label{equ:equ1}
\end{align}
where $\phi_{l}$ denotes the output of the $l$-th convolutional layer in a CNN, $\hat{Y}$ is the overall prediction, $w$ is the parameter of a pre-trained CNN and $\lambda_{l}$ is the trade-off parameter, controlling the influence of the loss in the $l$-th layer.
In this work, we use the first four convolutional layers of the VGG-16 model to calculate the SP loss between the ground-truth and the prediction.
The proposed loss ($\mathcal{L}_{wbce}+\mu\mathcal{L}_{sp}$) is continuously differentiable, so we can use the standard stochastic gradient descent (SGD) method to obtain the optimal parameters.

For saliency inference, we take the two paired images as input.
%
%
The saliency map is computed based on the output probabilities ($\textbf{s}_0$ and $\textbf{s}_1$) of each pixel with the softmax activation, which is denoted as:
\begin{align}
\text{Pr}(y_{j}=1|X;\theta) = \frac{exp(\textbf{s}_1)}{exp(\textbf{s}_0)+exp(\textbf{s}_1)}.
  \label{equ:equ8}
\end{align}
\vspace{-4mm}
\section{Experiments}
\vspace{-2mm}
In this section, we first introduce the experimental setups, including datasets and evaluation metrics.
Then we present the implementation details of our proposed approach.
Finally, we perform a series of experiments to thoroughly investigate the performance and impact of our proposed methods.
\vspace{-4mm}
\subsection{Experimental Setups}
{\flushleft\textbf{Datasets.}}
To train our model, we follow previous works~\cite{wang2016saliency,zhang2017amulet} and adopt the \textbf{MSRA10K}~\cite{borji2015salient} dataset, which has 10,000 training images with high quality pixel-wise saliency annotations.
Most of images in this dataset have a single salient object.
To make the model robust to the image translation variation and combat the over-fitting, we augment this dataset by random cropping and mirror reflection, producing 120,000 training image pairs totally.

For the performance evaluation, we evaluate the proposed method and compare our results with other state-of-the-art approaches on seven public datasets, described as follows:
\textbf{DUT-OMRON}~\cite{yang2013saliency} dataset has 5,168 high quality images.
Each image in this dataset has one or more objects with relatively complex backgrounds.
\textbf{DUTS-TE} dataset is the test set of currently largest saliency detection benchmark (DUTS)~\cite{wang2017learning}.
It contains 5,019 images with high quality pixel-wise annotations.
\textbf{ECSSD}~\cite{shi2016hierarchical} dataset contains 1,000 natural images, in which many semantically meaningful and complex structures are included.
\textbf{HKU-IS-TE}~\cite{li2015visual} dataset has 1,447 images with pixel-wise annotations.
Images of this dataset are well chosen to include multiple disconnected objects or objects touching the image boundary.
\textbf{PASCAL-S}~\cite{li2014secrets} dataset is generated from the PASCAL VOC~\cite{Everingham2010ThePV} dataset and contains 850 natural images with segmentation-based masks.
\textbf{SED}~\cite{borj2015salient} dataset has two non-overlapped subsets, \emph{i.e.}, SED1 and SED2.
SED1 has 100 images each containing only one salient object, while SED2 has 100 images each containing two salient objects.
\textbf{SOD}~\cite{jiang2013salient} dataset has 300 images, in which many images contain multiple objects either with low contrast or touching the image boundary.
{\flushleft\textbf{Evaluation Metrics.}}
To evaluate the performance of varied SOD algorithms, we adopt four metrics, including the widely used precision-recall (PR) curves, F-measure, mean absolute error (MAE)~\cite{borji2015salient} and recently proposed S-measure~\cite{fan2017structure}.
%
The PR curve of a specific dataset exhibits the mean precision and recall of saliency maps at different thresholds.
The F-measure is a weighted mean of average precision and average recall, calculated by
\begin{equation}
  F_{\eta} =\frac{(1+\eta^2)\times Precision\times Recall}{\eta^2\times Precision \times Recall}.
    \label{equ:equ9}
\end{equation}
%
We set $\eta^2$ to be 0.3 to weigh precision more than recall as suggested in~\cite{borji2015salient}.

%
%
For fair comparison on non-salient regions, we also calculate the mean absolute error (MAE) by
\begin{equation}
MAE = \frac{1}{W\times H}\sum_{x=1}^{W}\sum_{y=1}^{H}|S(x,y)-G(x,y)|,
  \label{equ:equ10}
\end{equation}
where $W$ and $H$ are the width and height of the input image. $S(x,y)$ and $G(x,y)$ are the pixel values of the saliency map and the binary ground truth at $(x,y)$, respectively.

To evaluate the spatial structure similarities of saliency maps, we also calculate the S-measure~\cite{fan2017structure}, defined as
\begin{equation}
S_{\lambda} = \lambda*S_{o}+(1-\lambda)*S_{r},
  \label{equ:equ11}
\end{equation}
where $\lambda \in [0,1]$ is the balance parameter, and is set as 0.5 typically. $S_{o}$ and $S_{r}$ are the object-aware and region-aware structural similarity, respectively.
\vspace{-4mm}
\subsection{Implementation Details}
The proposed model is implemented on the widely used deep learning framework, the Caffe toolbox~\cite{jia2014caffe}, with the MATLAB 2016 platform.
We train and test our methods in a quad-core PC machine with an NVIDIA TITAN 1070 GPU (with 8G memory) and an i5-6600 CPU.
Following~\cite{zhang2017amulet,zhang2017learning}, we perform training with the augmented training images from the MSRA10K dataset.
And we do not use any validation sets and train the model until its training loss converges.
The input image is uniformly resized into $384\times384\times3$ pixels and subtracted the ImageNet mean~\cite{imagenet_cvpr09}.
The weights of weight-stitching branches are initialized from the VGG-16 model~\cite{simonyan2014very}.
For the fusing branch, we initialize the weights by the ``msra'' method.
During the training, we use standard SGD method for updating the weights of the network, with a batch size 12, momentum 0.9 and weight decay 0.0005.
We set the base learning rate to 1e-8 and decrease the learning rate by 10\% when training loss reaches a flat.
In addition, we set $\mu=0.01$ to optimize the loss function for our experiments without further tuning.
The training process converges after 8 epoches.
When testing, our proposed SOD algorithm runs at about \textbf{6.7 fps}.
The source code will be made publicly available.
\vspace{-8mm}
\subsection{Comparisons with the State of the Art}
\begin{figure*}
\begin{center}
\resizebox{1\textwidth}{!}
{
\begin{tabular}{@{}c@{}c@{}c@{}c}
\includegraphics[width=0.24\linewidth,height=2.8cm]{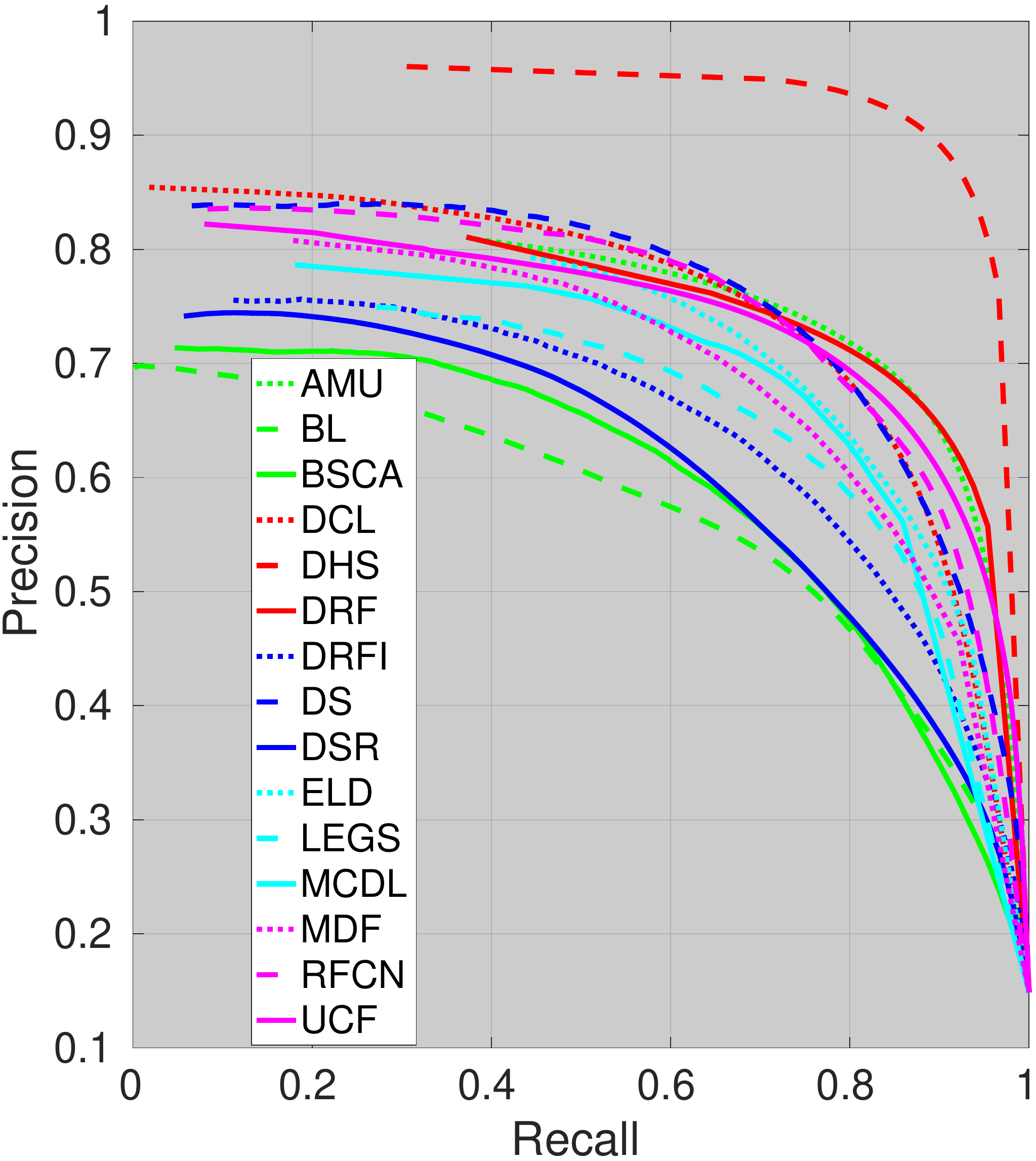} \ &
\includegraphics[width=0.24\linewidth,height=2.8cm]{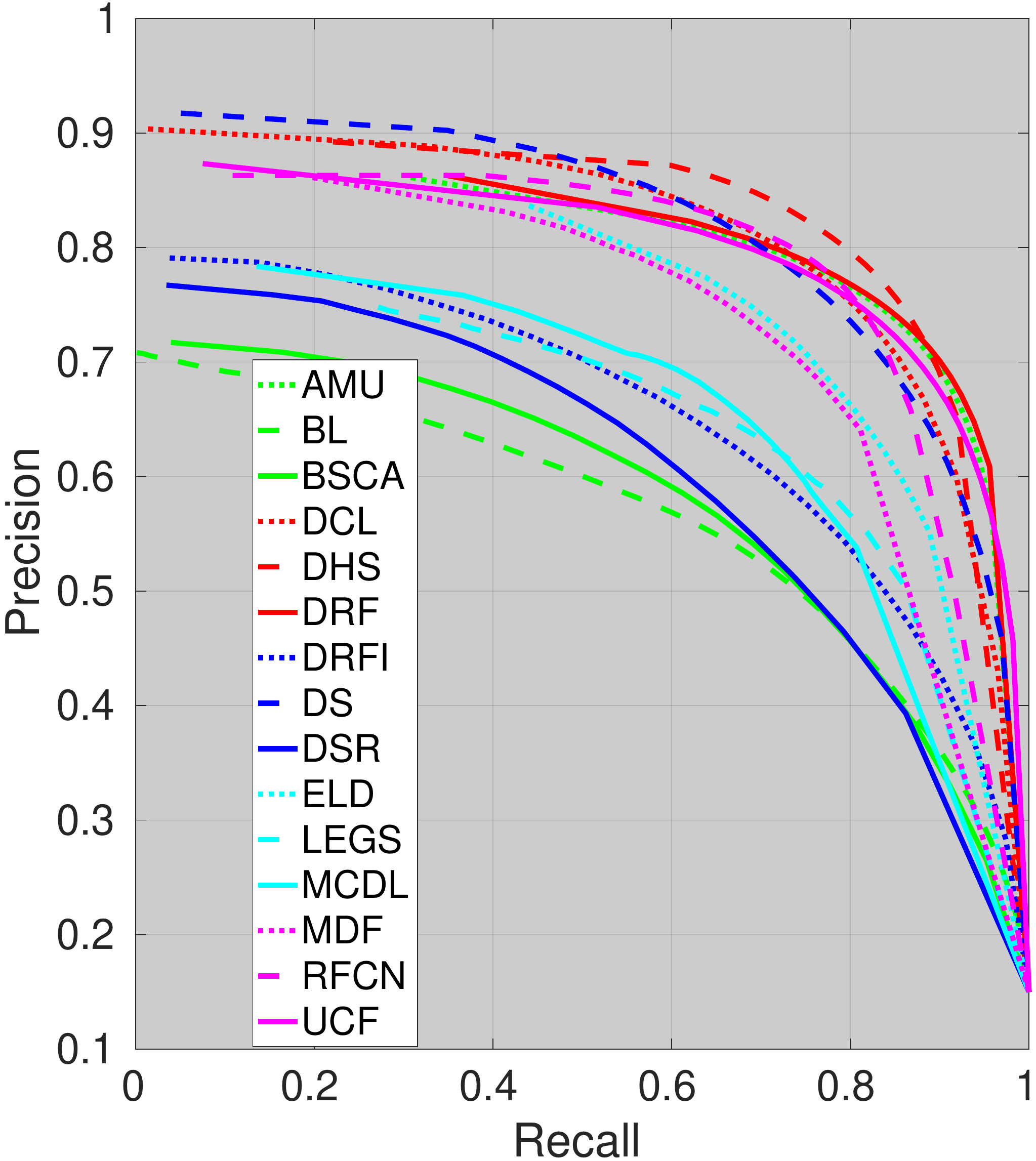} \ &
\includegraphics[width=0.24\linewidth,height=2.8cm]{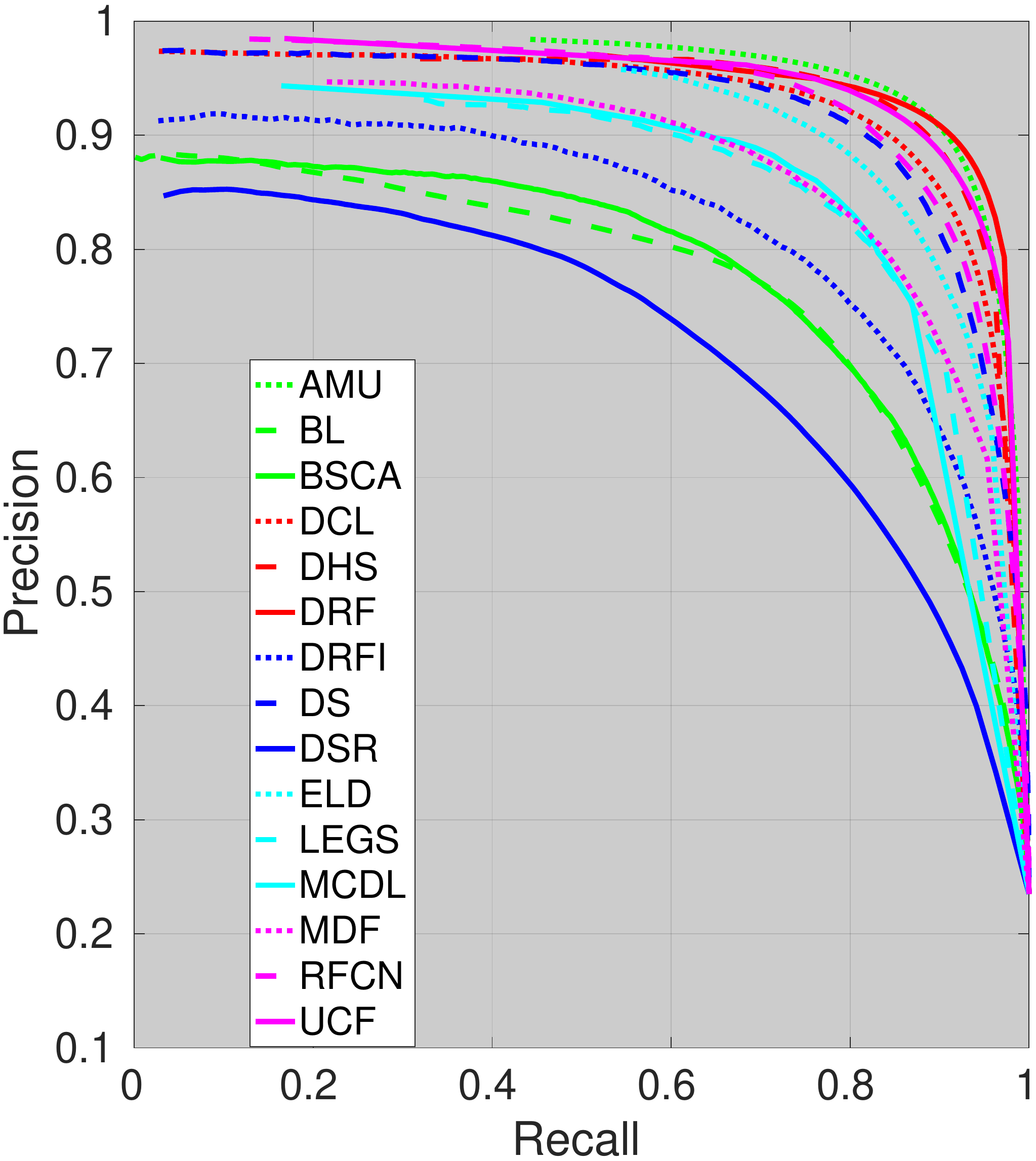} \ &
\includegraphics[width=0.24\linewidth,height=2.8cm]{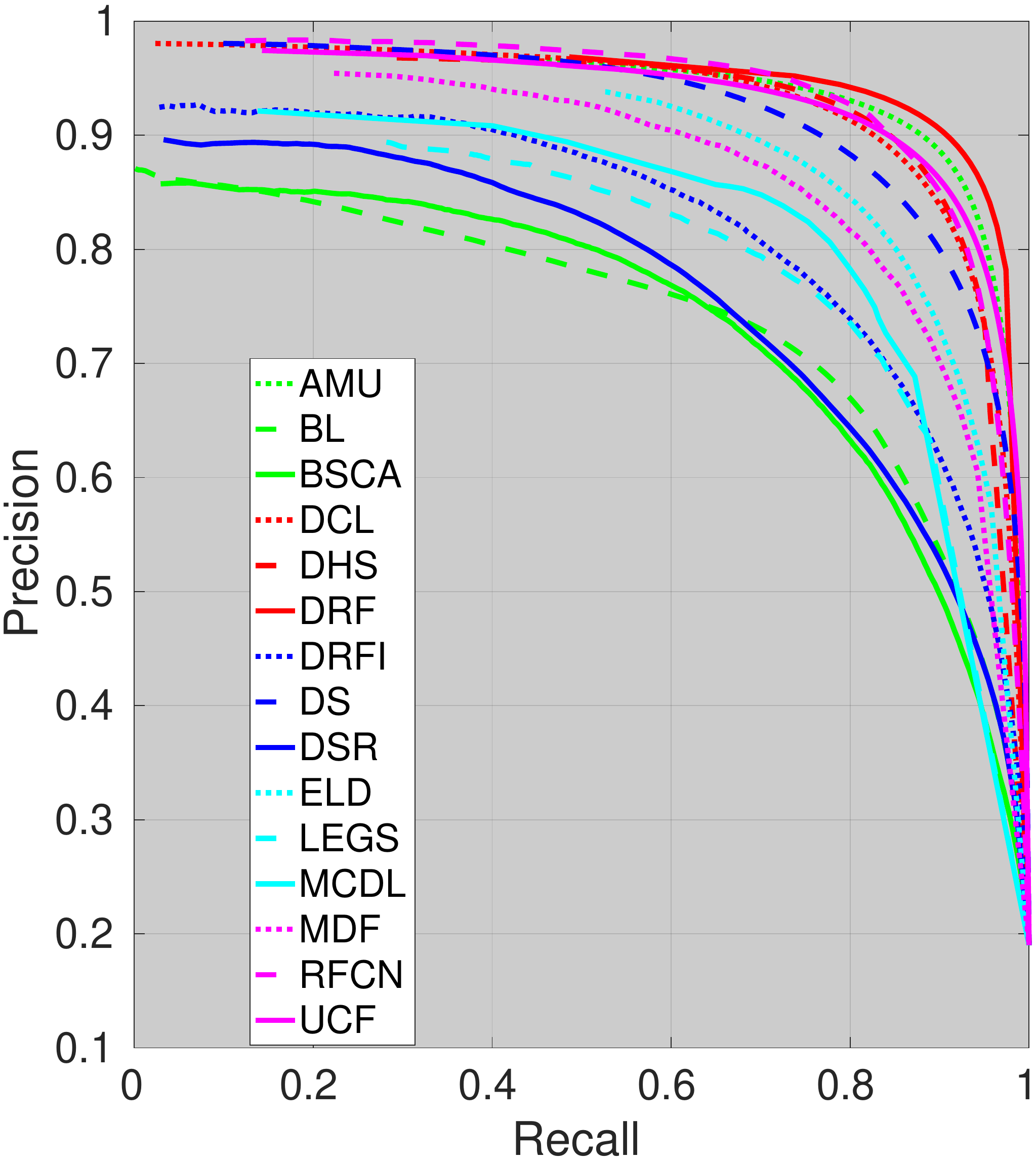} \ \\
 {\small(a) \textbf{DUT-OMRON}} & {\small(b) \textbf{DUTS-TE}} & {\small(c) \textbf{ECSSD}} & {\small(d) \textbf{HKU-IS}}\\
\includegraphics[width=0.24\linewidth,height=2.8cm]{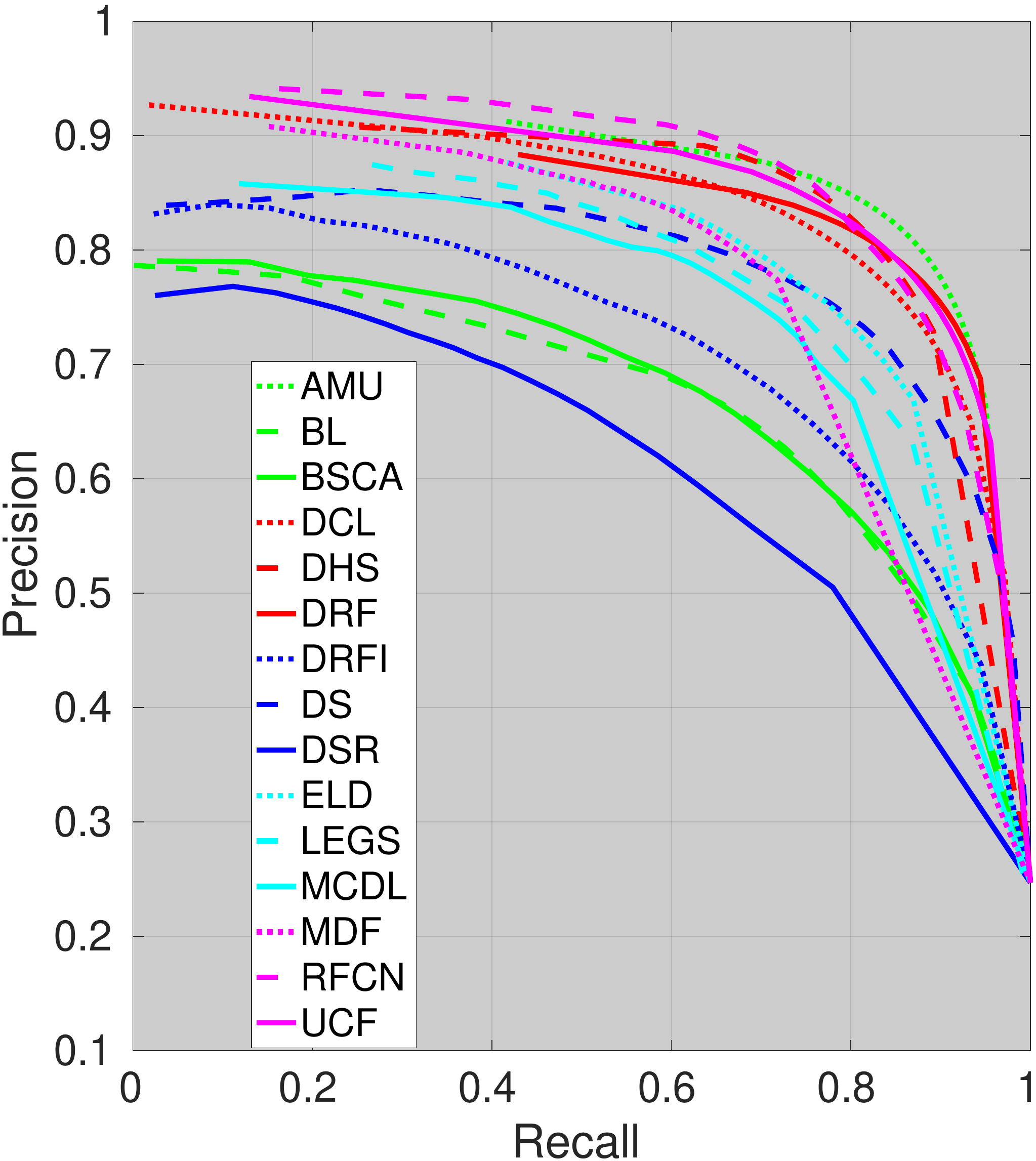} \ &
\includegraphics[width=0.24\linewidth,height=2.8cm]{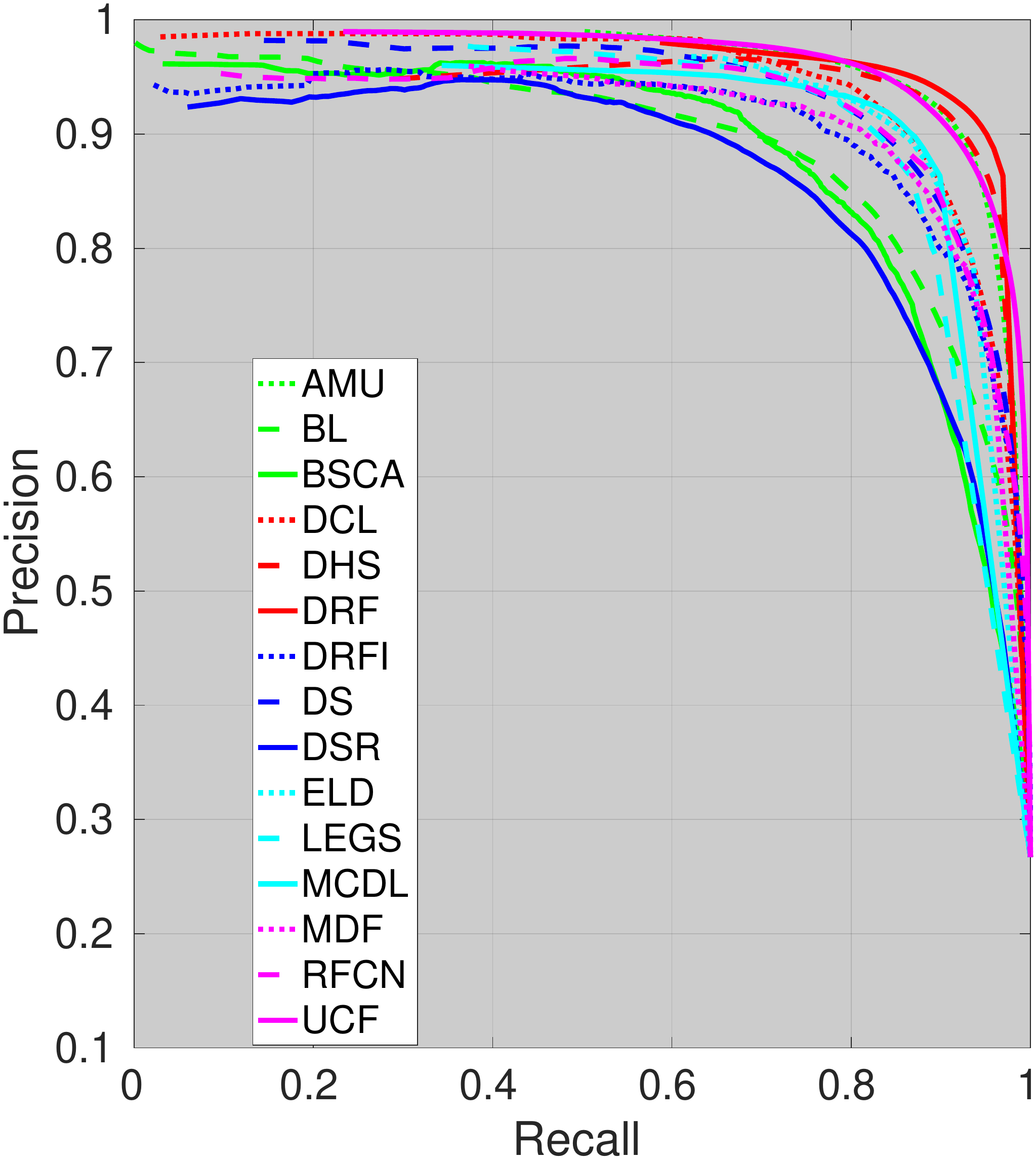} \ &
\includegraphics[width=0.24\linewidth,height=2.8cm]{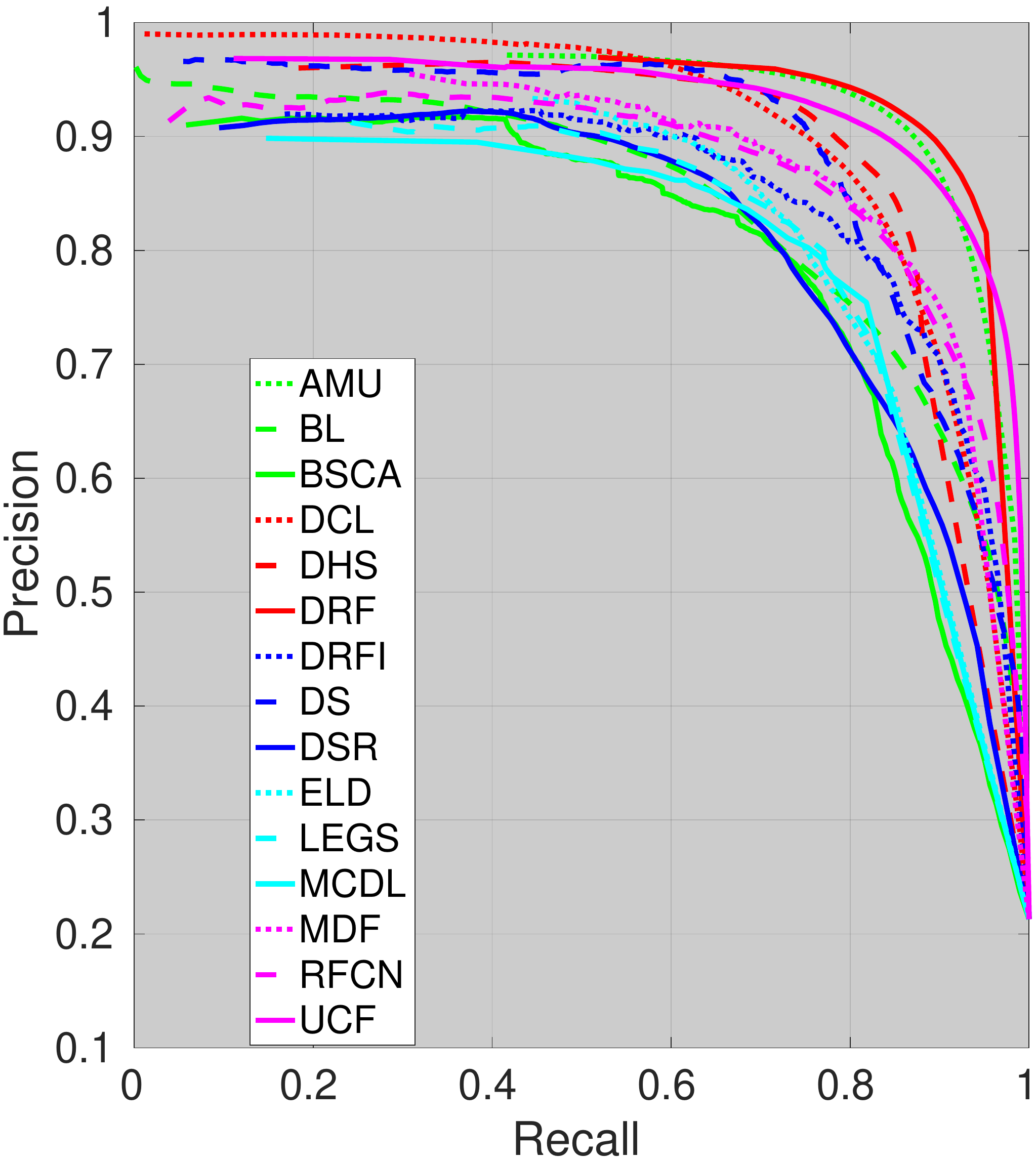} \ &
\includegraphics[width=0.24\linewidth,height=2.8cm]{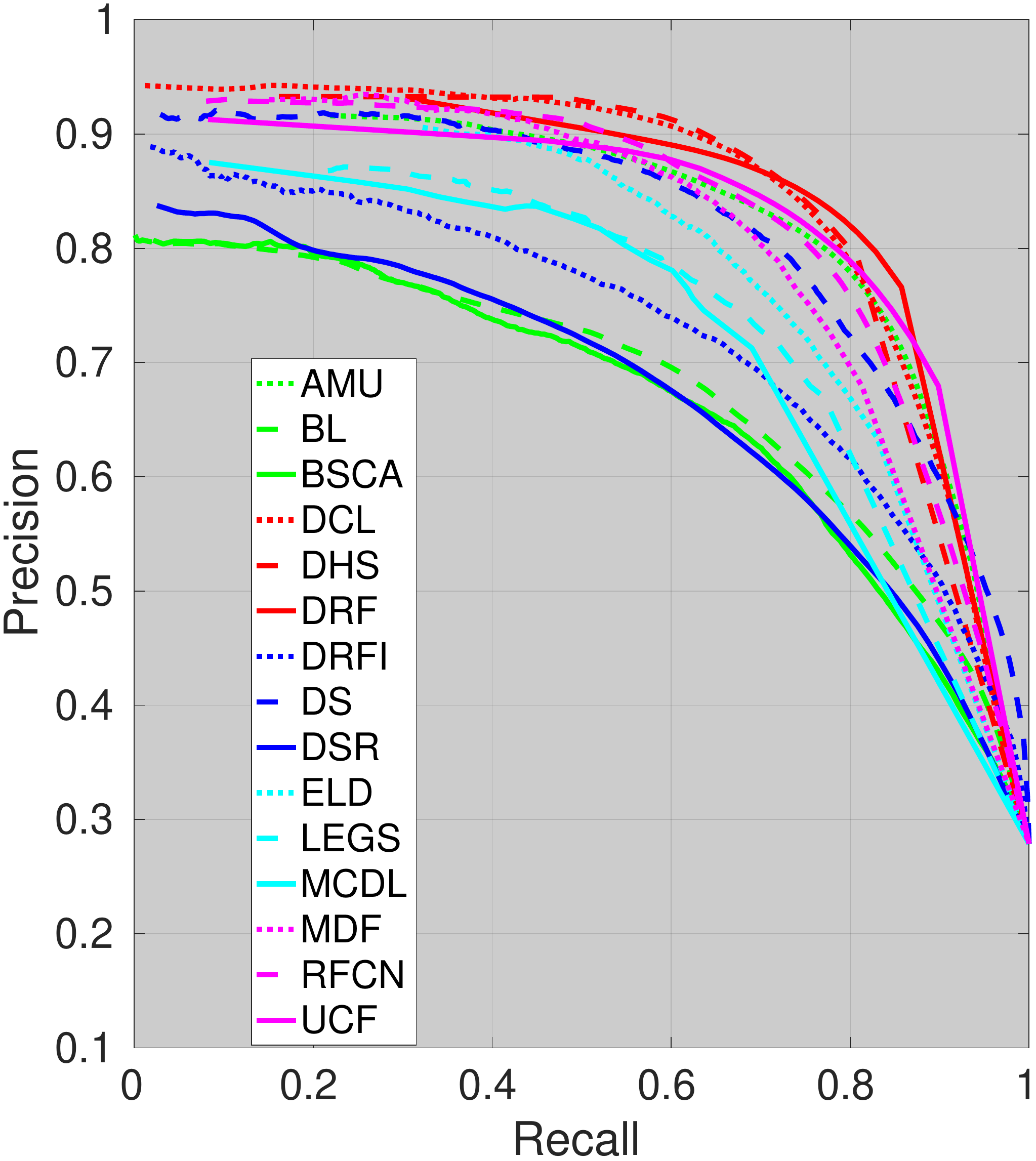} \ \\
 {\small(e) \textbf{PASCAL-S}} & {\small(f) \textbf{SED1}} & {\small(g) \textbf{SED2}} & {\small(h) \textbf{SOD}}\\
\\
\end{tabular}
}
\vspace{-6mm}
\caption{The PR curves of compared methods. Our method denotes as DRF.
\label{fig:PR-curve}}
\end{center}
\vspace{-10mm}
\end{figure*}

To fully evaluate the detection performance, we compare our proposed method with other 14 state-of-the-art ones, including 10 deep learning based algorithms (\textbf{AMU}~\cite{zhang2017amulet}, \textbf{DCL}~\cite{li2016dcl}, \textbf{DHS}~\cite{liu2016dhsnet}, \textbf{DS}~\cite{li2016ds}, \textbf{ELD}~\cite{lee2016deep}, \textbf{LEGS}~\cite{wang2015deep}, \textbf{MCDL}~\cite{zhao2015saliency}, \textbf{MDF}~\cite{li2015visual}, \textbf{RFCN}~\cite{wang2016saliency}, \textbf{UCF}~\cite{zhang2017learning}) and 4 outstanding conventional algorithms (\textbf{BL}~\cite{tong2015salient}, \textbf{BSCA}~\cite{qin2015saliency}, \textbf{DRFI}~\cite{jiang2013salient}, \textbf{DSR}~\cite{li2013saliency}).
For fair comparison, we use the detection results or original codes provided by authors with default setting.
And we report the results in Tab.~\ref{table:fauc1}-2 and Fig.~\ref{fig:PR-curve}-5.
\vspace{-2mm}
{\flushleft\textbf{Quantitative Results.}}
As illustrated in Tab.~\ref{table:fauc1}, Tab.~\ref{table:fauc2} and Fig.~\ref{fig:PR-curve}, our model achieves the best performance on most datasets.
Deep learning based methods achieve much better performance than traditional methods.
%
%
\begin{table*}
\begin{center}
\doublerulesep=0.4pt
\resizebox{1\textwidth}{!}
{
\begin{tabular}{|c|c|c|c|c|c|c|c|c|c|c|c|c|c|c|c|c|c|c|c|c|c|c|c|c|||c|c|c|c|c|c|c|c|||}
\hline
\multicolumn{4}{c|}{}
&\multicolumn{6}{c|}{\textbf{DUT-OMRON}}
&\multicolumn{6}{|c|}{\textbf{DUTS-TE}}
&\multicolumn{6}{|c|}{\textbf{ECSSD}}
&\multicolumn{6}{|c|}{\textbf{HKU-IS-TE}}
\\
\hline
\hline
\multicolumn{4}{c|}{Methods}
&\multicolumn{2}{|c|}{$F_\eta$}&\multicolumn{2}{|c|}{$MAE$}&\multicolumn{2}{|c|}{$S_\lambda$}
&\multicolumn{2}{|c|}{$F_\eta$}&\multicolumn{2}{|c|}{$MAE$}&\multicolumn{2}{|c|}{$S_\lambda$}
&\multicolumn{2}{|c|}{$F_\eta$}&\multicolumn{2}{|c|}{$MAE$}&\multicolumn{2}{|c|}{$S_\lambda$}
&\multicolumn{2}{|c}{$F_\eta$}&\multicolumn{2}{|c|}{$MAE$}&\multicolumn{2}{|c|}{$S_\lambda$}
\\
\hline
\hline
\multicolumn{4}{c|}{\textbf{Ours}}
&\multicolumn{2}{|c|}{\textcolor[rgb]{1,0,0}{0.701}}&\multicolumn{2}{|c|}{\textcolor[rgb]{1,0,0}{0.084}}&\multicolumn{2}{|c|}{\textcolor[rgb]{1,0,0}{0.784}}
&\multicolumn{2}{|c|}{\textcolor[rgb]{0,1,0}{0.722}}&\multicolumn{2}{|c|}{\textcolor[rgb]{0,1,0}{0.075}}&\multicolumn{2}{|c|}{\textcolor[rgb]{1,0,0}{0.812}}
&\multicolumn{2}{|c|}{\textcolor[rgb]{1,0,0}{0.886}}&\multicolumn{2}{|c|}{\textcolor[rgb]{1,0,0}{0.050}}&\multicolumn{2}{|c|}{\textcolor[rgb]{1,0,0}{0.903}}
&\multicolumn{2}{|c}{\textcolor[rgb]{1,0,0}{0.880}}&\multicolumn{2}{|c|}{\textcolor[rgb]{1,0,0}{0.037}}&\multicolumn{2}{|c|}{\textcolor[rgb]{1,0,0}{0.912}}
\\
\multicolumn{4}{c|}{\textbf{AMU}}
&\multicolumn{2}{|c|}{\textcolor[rgb]{0,0,1}{0.647}}&\multicolumn{2}{|c|}{0.098}&\multicolumn{2}{|c|}{\textcolor[rgb]{0,1,0}{0.771}}
&\multicolumn{2}{|c|}{0.682}&\multicolumn{2}{|c|}{\textcolor[rgb]{0,0,1}{0.085}}&\multicolumn{2}{|c|}{\textcolor[rgb]{0,0,1}{0.796}}
&\multicolumn{2}{|c|}{\textcolor[rgb]{0,0,1}{0.868}}&\multicolumn{2}{|c|}{\textcolor[rgb]{0,1,0}{0.059}}&\multicolumn{2}{|c|}{\textcolor[rgb]{0,1,0}{0.894}}
&\multicolumn{2}{|c}{0.843}&\multicolumn{2}{|c|}{\textcolor[rgb]{0,1,0}{0.050}}&\multicolumn{2}{|c|}{\textcolor[rgb]{0,1,0}{0.886}}
\\
\multicolumn{4}{c|}{\textbf{DCL}}
&\multicolumn{2}{|c|}{\textcolor[rgb]{0,1,0}{0.684}}&\multicolumn{2}{|c|}{0.157}&\multicolumn{2}{|c|}{0.743}
&\multicolumn{2}{|c|}{\textcolor[rgb]{0,0,1}{0.714}}&\multicolumn{2}{|c|}{0.150}&\multicolumn{2}{|c|}{0.785}
&\multicolumn{2}{|c|}{0.829}&\multicolumn{2}{|c|}{0.149}&\multicolumn{2}{|c|}{0.863}
&\multicolumn{2}{|c}{\textcolor[rgb]{0,0,1}{0.853}}&\multicolumn{2}{|c|}{0.136}&\multicolumn{2}{|c|}{0.859}
\\
\multicolumn{4}{c|}{\textbf{DHS}}
&\multicolumn{2}{|c|}{--}&\multicolumn{2}{|c|}{--}&\multicolumn{2}{|c|}{--}
&\multicolumn{2}{|c|}{\textcolor[rgb]{1,0,0}{0.724}}&\multicolumn{2}{|c|}{\textcolor[rgb]{1,0,0}{0.066}}&\multicolumn{2}{|c|}{\textcolor[rgb]{0,1,0}{0.809}}
&\multicolumn{2}{|c|}{\textcolor[rgb]{0,1,0}{0.872}}&\multicolumn{2}{|c|}{\textcolor[rgb]{0,0,1}{0.060}}&\multicolumn{2}{|c|}{0.884}
&\multicolumn{2}{|c}{\textcolor[rgb]{0,1,0}{0.854}}&\multicolumn{2}{|c|}{\textcolor[rgb]{0,0,1}{0.053}}&\multicolumn{2}{|c|}{0.869}
\\
\multicolumn{4}{c|}{\textbf{DS}}
&\multicolumn{2}{|c|}{0.603}&\multicolumn{2}{|c|}{0.120}&\multicolumn{2}{|c|}{0.741}
&\multicolumn{2}{|c|}{0.632}&\multicolumn{2}{|c|}{0.091}&\multicolumn{2}{|c|}{0.790}
&\multicolumn{2}{|c|}{0.826}&\multicolumn{2}{|c|}{0.122}&\multicolumn{2}{|c|}{0.821}
&\multicolumn{2}{|c}{0.787}&\multicolumn{2}{|c|}{0.077}&\multicolumn{2}{|c|}{0.854}
\\
\multicolumn{4}{c|}{\textbf{ELD}}
&\multicolumn{2}{|c|}{0.611}&\multicolumn{2}{|c|}{0.092}&\multicolumn{2}{|c|}{0.743}
&\multicolumn{2}{|c|}{0.628}&\multicolumn{2}{|c|}{0.098}&\multicolumn{2}{|c|}{0.749}
&\multicolumn{2}{|c|}{0.810}&\multicolumn{2}{|c|}{0.080}&\multicolumn{2}{|c|}{0.839}
&\multicolumn{2}{|c}{0.776}&\multicolumn{2}{|c|}{0.072}&\multicolumn{2}{|c|}{0.823}
\\
\multicolumn{4}{c|}{\textbf{LEGS}}
&\multicolumn{2}{|c|}{0.592}&\multicolumn{2}{|c|}{0.133}&\multicolumn{2}{|c|}{0.701}
&\multicolumn{2}{|c|}{0.585}&\multicolumn{2}{|c|}{0.138}&\multicolumn{2}{|c|}{0.687}
&\multicolumn{2}{|c|}{0.785}&\multicolumn{2}{|c|}{0.118}&\multicolumn{2}{|c|}{0.787}
&\multicolumn{2}{|c}{0.732}&\multicolumn{2}{|c|}{0.118}&\multicolumn{2}{|c|}{0.745}
\\
\multicolumn{4}{c|}{\textbf{MCDL}}
&\multicolumn{2}{|c|}{0.625}&\multicolumn{2}{|c|}{\textcolor[rgb]{0,1,0}{0.089}}&\multicolumn{2}{|c|}{0.739}
&\multicolumn{2}{|c|}{0.594}&\multicolumn{2}{|c|}{0.105}&\multicolumn{2}{|c|}{0.706}
&\multicolumn{2}{|c|}{0.796}&\multicolumn{2}{|c|}{0.101}&\multicolumn{2}{|c|}{0.803}
&\multicolumn{2}{|c}{0.760}&\multicolumn{2}{|c|}{0.091}&\multicolumn{2}{|c|}{0.786}
\\
\multicolumn{4}{c|}{\textbf{MDF}}
&\multicolumn{2}{|c|}{0.644}&\multicolumn{2}{|c|}{\textcolor[rgb]{0,0,1}{0.092}}&\multicolumn{2}{|c|}{0.703}
&\multicolumn{2}{|c|}{0.673}&\multicolumn{2}{|c|}{0.100}&\multicolumn{2}{|c|}{0.723}
&\multicolumn{2}{|c|}{0.807}&\multicolumn{2}{|c|}{0.105}&\multicolumn{2}{|c|}{0.776}
&\multicolumn{2}{|c}{0.802}&\multicolumn{2}{|c|}{0.095}&\multicolumn{2}{|c|}{0.779}
\\
\multicolumn{4}{c|}{\textbf{RFCN}}
&\multicolumn{2}{|c|}{0.627}&\multicolumn{2}{|c|}{0.111}&\multicolumn{2}{|c|}{\textcolor[rgb]{0,0,1}{0.752}}
&\multicolumn{2}{|c|}{0.712}&\multicolumn{2}{|c|}{0.090}&\multicolumn{2}{|c|}{0.784}
&\multicolumn{2}{|c|}{0.834}&\multicolumn{2}{|c|}{0.107}&\multicolumn{2}{|c|}{0.852}
&\multicolumn{2}{|c}{0.838}&\multicolumn{2}{|c|}{0.088}&\multicolumn{2}{|c|}{0.860}
\\
\multicolumn{4}{c|}{\textbf{UCF}}
&\multicolumn{2}{|c|}{0.621}&\multicolumn{2}{|c|}{0.120}&\multicolumn{2}{|c|}{0.748}
&\multicolumn{2}{|c|}{0.635}&\multicolumn{2}{|c|}{0.112}&\multicolumn{2}{|c|}{0.777}
&\multicolumn{2}{|c|}{0.844}&\multicolumn{2}{|c|}{0.069}&\multicolumn{2}{|c|}{\textcolor[rgb]{0,0,1}{0.884}}
&\multicolumn{2}{|c}{0.823}&\multicolumn{2}{|c|}{0.061}&\multicolumn{2}{|c|}{\textcolor[rgb]{0,0,1}{0.874}}
\\
\hline
\hline
\multicolumn{4}{c|}{\textbf{BL}}
&\multicolumn{2}{|c|}{0.499}&\multicolumn{2}{|c|}{0.239}&\multicolumn{2}{|c|}{0.625}
&\multicolumn{2}{|c|}{0.490}&\multicolumn{2}{|c|}{0.238}&\multicolumn{2}{|c|}{0.615}
&\multicolumn{2}{|c|}{0.684}&\multicolumn{2}{|c|}{0.216}&\multicolumn{2}{|c|}{0.714}
&\multicolumn{2}{|c}{0.666}&\multicolumn{2}{|c|}{0.207}&\multicolumn{2}{|c|}{0.702}
\\
\multicolumn{4}{c|}{\textbf{BSCA}}
&\multicolumn{2}{|c|}{0.509}&\multicolumn{2}{|c|}{0.190}&\multicolumn{2}{|c|}{0.652}
&\multicolumn{2}{|c|}{0.500}&\multicolumn{2}{|c|}{0.196}&\multicolumn{2}{|c|}{0.633}
&\multicolumn{2}{|c|}{0.705}&\multicolumn{2}{|c|}{0.182}&\multicolumn{2}{|c|}{0.725}
&\multicolumn{2}{|c}{0.658}&\multicolumn{2}{|c|}{0.175}&\multicolumn{2}{|c|}{0.705}
\\
\multicolumn{4}{c|}{\textbf{DRFI}}
&\multicolumn{2}{|c|}{0.550}&\multicolumn{2}{|c|}{0.138}&\multicolumn{2}{|c|}{0.688}
&\multicolumn{2}{|c|}{0.541}&\multicolumn{2}{|c|}{0.175}&\multicolumn{2}{|c|}{0.662}
&\multicolumn{2}{|c|}{0.733}&\multicolumn{2}{|c|}{0.164}&\multicolumn{2}{|c|}{0.752}
&\multicolumn{2}{|c}{0.726}&\multicolumn{2}{|c|}{0.145}&\multicolumn{2}{|c|}{0.743}
\\
\multicolumn{4}{c|}{\textbf{DSR}}
&\multicolumn{2}{|c|}{0.524}&\multicolumn{2}{|c|}{0.139}&\multicolumn{2}{|c|}{0.660}
&\multicolumn{2}{|c|}{0.518}&\multicolumn{2}{|c|}{0.145}&\multicolumn{2}{|c|}{0.646}
&\multicolumn{2}{|c|}{0.662}&\multicolumn{2}{|c|}{0.178}&\multicolumn{2}{|c|}{0.731}
&\multicolumn{2}{|c}{0.682}&\multicolumn{2}{|c|}{0.142}&\multicolumn{2}{|c|}{0.701}
\\
\hline
\end{tabular}
}
\vspace{1mm}
\caption{Quantitative comparison with 15 methods on 4 large-scale datasets. The best three results are shown in \textcolor[rgb]{1,0,0}{red},~\textcolor[rgb]{0,1,0}{green} and \textcolor[rgb]{0,0,1}{blue}, respectively. ``--'' means corresponding methods are trained on that dataset. Our method ranks first or second. }
\label{table:fauc1}
\end{center}
\vspace{-6mm}
\end{table*}
\begin{table*}
\begin{center}
\doublerulesep=0.4pt
\resizebox{1\textwidth}{!}
{
\begin{tabular}{|c|c|c|c|c|c|c|c|c|c|c|c|c|c|c|c|c|c|c|c|c|c|c|c|c|||c|c|c|c|c|c|c|c|||}
\hline
\multicolumn{4}{c|}{}
&\multicolumn{6}{|c|}{\textbf{PASCAL-S}}
&\multicolumn{6}{|c|}{\textbf{SED1}}
&\multicolumn{6}{|c|}{\textbf{SED2}}
&\multicolumn{6}{|c|}{\textbf{SOD}}
\\
\hline
\hline
\multicolumn{4}{c|}{Methods}
&\multicolumn{2}{|c|}{$F_\eta$}&\multicolumn{2}{|c|}{$MAE$}&\multicolumn{2}{|c|}{$S_\lambda$}
&\multicolumn{2}{|c|}{$F_\eta$}&\multicolumn{2}{|c|}{$MAE$}&\multicolumn{2}{|c|}{$S_\lambda$}
&\multicolumn{2}{|c|}{$F_\eta$}&\multicolumn{2}{|c|}{$MAE$}&\multicolumn{2}{|c|}{$S_\lambda$}
&\multicolumn{2}{|c|}{$F_\eta$}&\multicolumn{2}{|c|}{$MAE$}&\multicolumn{2}{|c|}{$S_\lambda$}
\\
\hline
\hline
\multicolumn{4}{c|}{\textbf{Ours}}
&\multicolumn{2}{|c|}{\textcolor[rgb]{1,0,0}{0.784}}&\multicolumn{2}{|c|}{\textcolor[rgb]{0,0,1}{0.100}}&\multicolumn{2}{|c|}{\textcolor[rgb]{0,1,0}{0.813}}
&\multicolumn{2}{|c|}{\textcolor[rgb]{1,0,0}{0.921}}&\multicolumn{2}{|c|}{\textcolor[rgb]{1,0,0}{0.045}}&\multicolumn{2}{|c|}{\textcolor[rgb]{1,0,0}{0.911}}
&\multicolumn{2}{|c|}{\textcolor[rgb]{1,0,0}{0.875}}&\multicolumn{2}{|c|}{\textcolor[rgb]{1,0,0}{0.046}}&\multicolumn{2}{|c|}{\textcolor[rgb]{1,0,0}{0.874}}
&\multicolumn{2}{|c|}{\textcolor[rgb]{1,0,0}{0.793}}&\multicolumn{2}{|c|}{\textcolor[rgb]{1,0,0}{0.121}}&\multicolumn{2}{|c|}{\textcolor[rgb]{1,0,0}{0.778}}
\\
\multicolumn{4}{c|}{\textbf{AMU}}
&\multicolumn{2}{|c|}{\textcolor[rgb]{0,0,1}{0.768}}&\multicolumn{2}{|c|}{\textcolor[rgb]{0,1,0}{0.098}}&\multicolumn{2}{|c|}{\textcolor[rgb]{1,0,0}{0.820}}
&\multicolumn{2}{|c|}{\textcolor[rgb]{0,1,0}{0.892}}&\multicolumn{2}{|c|}{\textcolor[rgb]{0,0,1}{0.060}}&\multicolumn{2}{|c|}{0.893}
&\multicolumn{2}{|c|}{\textcolor[rgb]{0,1,0}{0.830}}&\multicolumn{2}{|c|}{\textcolor[rgb]{0,1,0}{0.062}}&\multicolumn{2}{|c|}{\textcolor[rgb]{0,1,0}{0.852}}
&\multicolumn{2}{|c|}{\textcolor[rgb]{0,0,1}{0.745}}&\multicolumn{2}{|c|}{\textcolor[rgb]{0,0,1}{0.144}}&\multicolumn{2}{|c|}{\textcolor[rgb]{0,0,1}{0.753}}
\\
\multicolumn{4}{c|}{\textbf{DCL}}
&\multicolumn{2}{|c|}{0.714}&\multicolumn{2}{|c|}{0.181}&\multicolumn{2}{|c|}{0.791}
&\multicolumn{2}{|c|}{0.855}&\multicolumn{2}{|c|}{0.151}&\multicolumn{2}{|c|}{0.845}
&\multicolumn{2}{|c|}{0.795}&\multicolumn{2}{|c|}{0.157}&\multicolumn{2}{|c|}{0.760}
&\multicolumn{2}{|c|}{0.741}&\multicolumn{2}{|c|}{0.194}&\multicolumn{2}{|c|}{0.748}
\\
\multicolumn{4}{c|}{\textbf{DHS}}
&\multicolumn{2}{|c|}{\textcolor[rgb]{0,1,0}{0.777}}&\multicolumn{2}{|c|}{\textcolor[rgb]{1,0,0}{0.095}}&\multicolumn{2}{|c|}{\textcolor[rgb]{0,0,1}{0.807}}
&\multicolumn{2}{|c|}{\textcolor[rgb]{0,0,1}{0.888}}&\multicolumn{2}{|c|}{\textcolor[rgb]{0,1,0}{0.055}}&\multicolumn{2}{|c|}{\textcolor[rgb]{0,0,1}{0.894}}
&\multicolumn{2}{|c|}{\textcolor[rgb]{0,0,1}{0.822}}&\multicolumn{2}{|c|}{0.080}&\multicolumn{2}{|c|}{0.796}
&\multicolumn{2}{|c|}{\textcolor[rgb]{0,1,0}{0.775}}&\multicolumn{2}{|c|}{\textcolor[rgb]{0,1,0}{0.129}}&\multicolumn{2}{|c|}{0.750}
\\
\multicolumn{4}{c|}{\textbf{DS}}
&\multicolumn{2}{|c|}{0.659}&\multicolumn{2}{|c|}{0.176}&\multicolumn{2}{|c|}{0.739}
&\multicolumn{2}{|c|}{0.845}&\multicolumn{2}{|c|}{0.093}&\multicolumn{2}{|c|}{0.859}
&\multicolumn{2}{|c|}{0.754}&\multicolumn{2}{|c|}{0.123}&\multicolumn{2}{|c|}{0.776}
&\multicolumn{2}{|c|}{0.698}&\multicolumn{2}{|c|}{0.189}&\multicolumn{2}{|c|}{0.712}
\\
\multicolumn{4}{c|}{\textbf{ELD}}
&\multicolumn{2}{|c|}{0.718}&\multicolumn{2}{|c|}{0.123}&\multicolumn{2}{|c|}{0.757}
&\multicolumn{2}{|c|}{0.872}&\multicolumn{2}{|c|}{0.067}&\multicolumn{2}{|c|}{0.864}
&\multicolumn{2}{|c|}{0.759}&\multicolumn{2}{|c|}{0.103}&\multicolumn{2}{|c|}{0.769}
&\multicolumn{2}{|c|}{0.712}&\multicolumn{2}{|c|}{0.155}&\multicolumn{2}{|c|}{0.705}
\\
\multicolumn{4}{c|}{\textbf{LEGS}}
&\multicolumn{2}{|c|}{--}&\multicolumn{2}{|c|}{--}&\multicolumn{2}{|c|}{--}
&\multicolumn{2}{|c|}{0.854}&\multicolumn{2}{|c|}{0.103}&\multicolumn{2}{|c|}{0.828}
&\multicolumn{2}{|c|}{0.736}&\multicolumn{2}{|c|}{0.124}&\multicolumn{2}{|c|}{0.716}
&\multicolumn{2}{|c|}{0.683}&\multicolumn{2}{|c|}{0.196}&\multicolumn{2}{|c|}{0.657}
\\
\multicolumn{4}{c|}{\textbf{MCDL}}
&\multicolumn{2}{|c|}{0.691}&\multicolumn{2}{|c|}{0.145}&\multicolumn{2}{|c|}{0.719}
&\multicolumn{2}{|c|}{0.878}&\multicolumn{2}{|c|}{0.077}&\multicolumn{2}{|c|}{0.855}
&\multicolumn{2}{|c|}{0.757}&\multicolumn{2}{|c|}{0.116}&\multicolumn{2}{|c|}{0.742}
&\multicolumn{2}{|c|}{0.677}&\multicolumn{2}{|c|}{0.181}&\multicolumn{2}{|c|}{0.650}
\\
\multicolumn{4}{c|}{\textbf{MDF}}
&\multicolumn{2}{|c|}{0.709}&\multicolumn{2}{|c|}{0.146}&\multicolumn{2}{|c|}{0.692}
&\multicolumn{2}{|c|}{0.842}&\multicolumn{2}{|c|}{0.099}&\multicolumn{2}{|c|}{0.833}
&\multicolumn{2}{|c|}{0.800}&\multicolumn{2}{|c|}{0.101}&\multicolumn{2}{|c|}{0.772}
&\multicolumn{2}{|c|}{0.721}&\multicolumn{2}{|c|}{0.165}&\multicolumn{2}{|c|}{0.674}
\\
\multicolumn{4}{c|}{\textbf{RFCN}}
&\multicolumn{2}{|c|}{0.751}&\multicolumn{2}{|c|}{0.132}&\multicolumn{2}{|c|}{0.799}
&\multicolumn{2}{|c|}{0.850}&\multicolumn{2}{|c|}{0.117}&\multicolumn{2}{|c|}{0.832}
&\multicolumn{2}{|c|}{0.767}&\multicolumn{2}{|c|}{0.113}&\multicolumn{2}{|c|}{0.784}
&\multicolumn{2}{|c|}{0.743}&\multicolumn{2}{|c|}{0.170}&\multicolumn{2}{|c|}{0.730}
\\
\multicolumn{4}{c|}{\textbf{UCF}}
&\multicolumn{2}{|c|}{0.735}&\multicolumn{2}{|c|}{0.115}&\multicolumn{2}{|c|}{0.806}
&\multicolumn{2}{|c|}{0.865}&\multicolumn{2}{|c|}{0.063}&\multicolumn{2}{|c|}{\textcolor[rgb]{0,1,0}{0.896}}
&\multicolumn{2}{|c|}{0.810}&\multicolumn{2}{|c|}{\textcolor[rgb]{0,0,1}{0.068}}&\multicolumn{2}{|c|}{\textcolor[rgb]{0,0,1}{0.846}}
&\multicolumn{2}{|c|}{0.738}&\multicolumn{2}{|c|}{0.148}&\multicolumn{2}{|c|}{\textcolor[rgb]{0,1,0}{0.762}}
\\
\hline
\hline
\multicolumn{4}{c|}{\textbf{BL}}
&\multicolumn{2}{|c|}{0.574}&\multicolumn{2}{|c|}{0.249}&\multicolumn{2}{|c|}{0.647}
&\multicolumn{2}{|c|}{0.780}&\multicolumn{2}{|c|}{0.185}&\multicolumn{2}{|c|}{0.783}
&\multicolumn{2}{|c|}{0.713}&\multicolumn{2}{|c|}{0.186}&\multicolumn{2}{|c|}{0.705}
&\multicolumn{2}{|c|}{0.580}&\multicolumn{2}{|c|}{0.267}&\multicolumn{2}{|c|}{0.625}
\\
\multicolumn{4}{c|}{\textbf{BSCA}}
&\multicolumn{2}{|c|}{0.601}&\multicolumn{2}{|c|}{0.223}&\multicolumn{2}{|c|}{0.652}
&\multicolumn{2}{|c|}{0.805}&\multicolumn{2}{|c|}{0.153}&\multicolumn{2}{|c|}{0.785}
&\multicolumn{2}{|c|}{0.706}&\multicolumn{2}{|c|}{0.158}&\multicolumn{2}{|c|}{0.714}
&\multicolumn{2}{|c|}{0.584}&\multicolumn{2}{|c|}{0.252}&\multicolumn{2}{|c|}{0.621}
\\
\multicolumn{4}{c|}{\textbf{DRFI}}
&\multicolumn{2}{|c|}{0.618}&\multicolumn{2}{|c|}{0.207}&\multicolumn{2}{|c|}{0.670}
&\multicolumn{2}{|c|}{0.807}&\multicolumn{2}{|c|}{0.148}&\multicolumn{2}{|c|}{0.797}
&\multicolumn{2}{|c|}{0.745}&\multicolumn{2}{|c|}{0.133}&\multicolumn{2}{|c|}{0.750}
&\multicolumn{2}{|c|}{0.634}&\multicolumn{2}{|c|}{0.224}&\multicolumn{2}{|c|}{0.624}
\\
\multicolumn{4}{c|}{\textbf{DSR}}
&\multicolumn{2}{|c|}{0.558}&\multicolumn{2}{|c|}{0.215}&\multicolumn{2}{|c|}{0.594}
&\multicolumn{2}{|c|}{0.791}&\multicolumn{2}{|c|}{0.158}&\multicolumn{2}{|c|}{0.736}
&\multicolumn{2}{|c|}{0.712}&\multicolumn{2}{|c|}{0.141}&\multicolumn{2}{|c|}{0.715}
&\multicolumn{2}{|c|}{0.596}&\multicolumn{2}{|c|}{0.234}&\multicolumn{2}{|c|}{0.596}
\\
\hline
\end{tabular}
}
\vspace{2mm}
\caption{Quantitative comparison with 15 methods on 4 complex scene image datasets. The best three results are shown in \textcolor[rgb]{1,0,0}{red},~\textcolor[rgb]{0,1,0}{green} and \textcolor[rgb]{0,0,1}{blue}, respectively. ``--'' means corresponding methods are trained on that dataset. Our method ranks first or second.}
\label{table:fauc2}
\end{center}
\vspace{-10mm}
\end{table*}
\begin{table*}
\begin{center}
\resizebox{1\textwidth}{!}
{
\begin{tabular}{|c|c|c|c|c|c|c|}
\hline
Models &(a) \small{ICNN}-hf+$\mathcal{L}_{bce}$&(b) \small{ICNN}+$\mathcal{L}_{bce}$&(c) \small{ICNN}+$\mathcal{L}_{wbce}$&(d) \small{ICNN}+$\mathcal{L}_{bce}$+$\mathcal{L}_{sp}$&The proposed \\
\hline
$F_\eta$       &0.832&0.854& \textcolor[rgb]{0,1,0}{0.876}& \textcolor[rgb]{0,0,1}{0.871}&\textcolor[rgb]{1,0,0}{0.886}        \\
\hline
$MAE$          &0.098&0.076& \textcolor[rgb]{0,1,0}{0.054}& \textcolor[rgb]{0,0,1}{0.068}&\textcolor[rgb]{1,0,0}{0.050}       \\
\hline
$S_\lambda$    &0.845&0.862& \textcolor[rgb]{0,0,1}{0.871}& \textcolor[rgb]{0,1,0}{0.886}&\textcolor[rgb]{1,0,0}{0.903}       \\
\hline
\end{tabular}
}
\end{center}
\caption{Results with different loss functions on the ECSSD dataset. The best three results are shown in \textcolor[rgb]{1,0,0}{red},~\textcolor[rgb]{0,1,0}{green} and \textcolor[rgb]{0,0,1}{blue}, respectively.}
\label{table:loss}
\vspace{-6mm}
\end{table*}
From these results, we have other notable observations: (1) Compared with the existing state-of-the-art methods, our method outperforms other competing ones (except DHS) with a large margin on the four large-scale datasets, especially on DUT-OMRON, ECSSD and HKU-IS-TE.
%
%
(2) Our method achieves higher S-measure on complex scene datasets, \emph{e.g.}, the DUT-OMRON, SED and SOD datasets.
We attribute this result to our image separation method.
(3) Without segmentation pre-training and any post-processing, such as CRF or superpixel refinement, our method still achieves better results than DCL, ELD, MCDL and RFCN, especially on the HKU-IS, SED and SOD datasets.
In average, our method achieves about 4\% performance leap of F-measure and around 2\% improvement of S-measure, as well as around 4\% decrease in MAE compared with existing best methods.
(4) Compared to the top-ranked methods, \emph{i.e.}, AMU and DHS, our method is inferior on the DUTS-TE and PASCAL-S datasets under several metrics.
However, our method ranks at the second place and is still very comparable.
\begin{figure*}
\centering
\resizebox{1\textwidth}{!}
{
\begin{tabular}{@{}c@{}c@{}c@{}c@{}c@{}c@{}c@{}c@{}c@{}c@{}c@{}c}
\vspace{-0.5mm}
\includegraphics[width=0.085\linewidth,height=1.2cm]{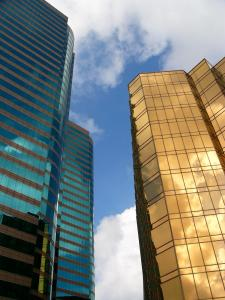}\ &
\includegraphics[width=0.085\linewidth,height=1.2cm]{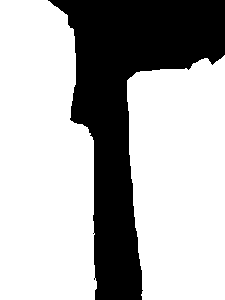}\ &
\includegraphics[width=0.085\linewidth,height=1.2cm]{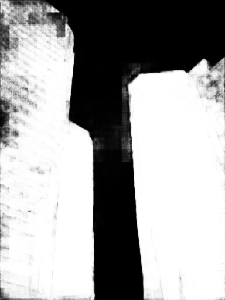}\ &
\includegraphics[width=0.085\linewidth,height=1.2cm]{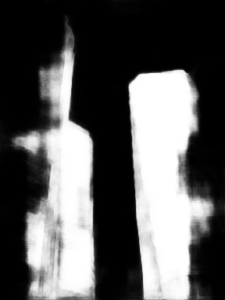}\ &
\includegraphics[width=0.085\linewidth,height=1.2cm]{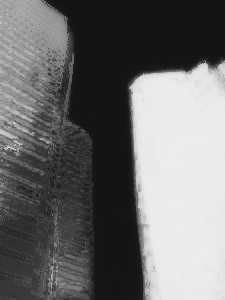}\ &
\includegraphics[width=0.085\linewidth,height=1.2cm]{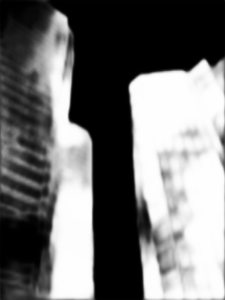}\ &
\includegraphics[width=0.085\linewidth,height=1.2cm]{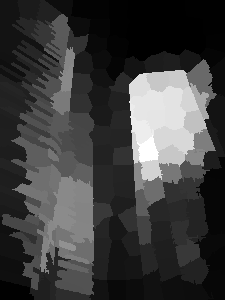}\ &
\includegraphics[width=0.085\linewidth,height=1.2cm]{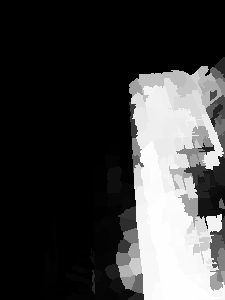}\ &
\includegraphics[width=0.085\linewidth,height=1.2cm]{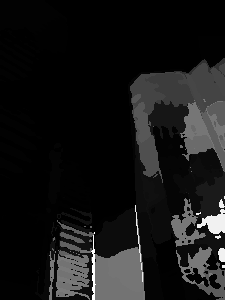}\ &
\includegraphics[width=0.085\linewidth,height=1.2cm]{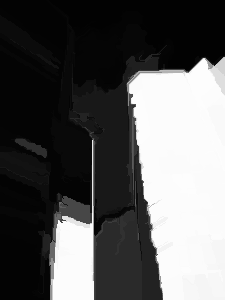}\ &
\includegraphics[width=0.085\linewidth,height=1.2cm]{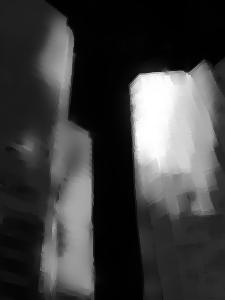}\ &
\includegraphics[width=0.085\linewidth,height=1.2cm]{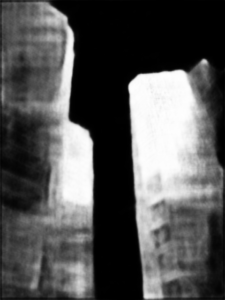}\ \\
\vspace{-0.5mm}
\includegraphics[width=0.085\linewidth,height=1.2cm]{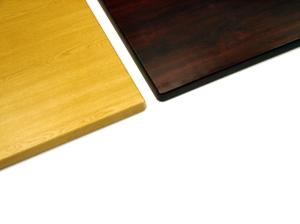}\ &
\includegraphics[width=0.085\linewidth,height=1.2cm]{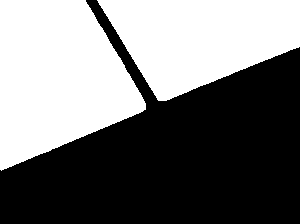}\ &
\includegraphics[width=0.085\linewidth,height=1.2cm]{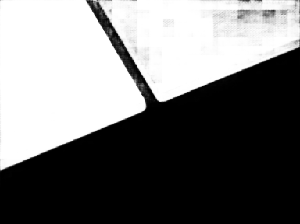}\ &
\includegraphics[width=0.085\linewidth,height=1.2cm]{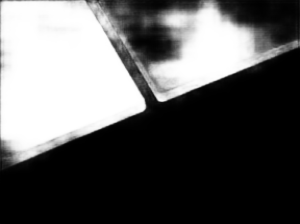}\ &
\includegraphics[width=0.085\linewidth,height=1.2cm]{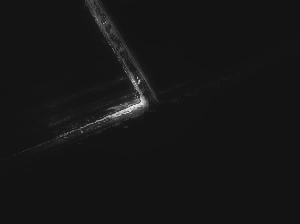}\ &
\includegraphics[width=0.085\linewidth,height=1.2cm]{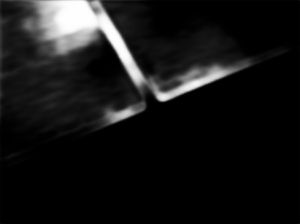}\ &
\includegraphics[width=0.085\linewidth,height=1.2cm]{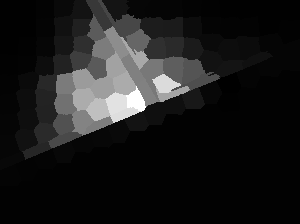}\ &
\includegraphics[width=0.085\linewidth,height=1.2cm]{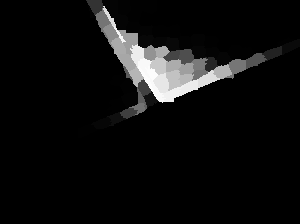}\ &
\includegraphics[width=0.085\linewidth,height=1.2cm]{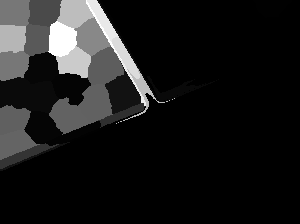}\ &
\includegraphics[width=0.085\linewidth,height=1.2cm]{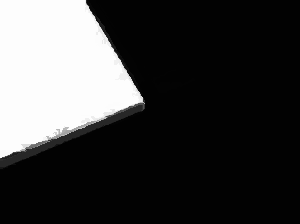}\ &
\includegraphics[width=0.085\linewidth,height=1.2cm]{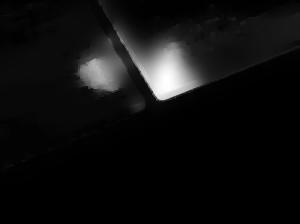}\ &
\includegraphics[width=0.085\linewidth,height=1.2cm]{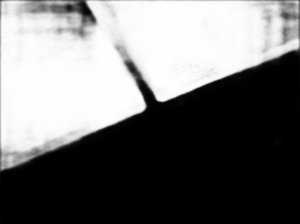}\ \\
\vspace{-0.5mm}
\includegraphics[width=0.085\linewidth,height=1.25cm]{0070.jpg}\ &
\includegraphics[width=0.085\linewidth,height=1.25cm]{0070_GT.png}\ &
\includegraphics[width=0.085\linewidth,height=1.25cm]{0070_DRF.png}\ &
\includegraphics[width=0.085\linewidth,height=1.25cm]{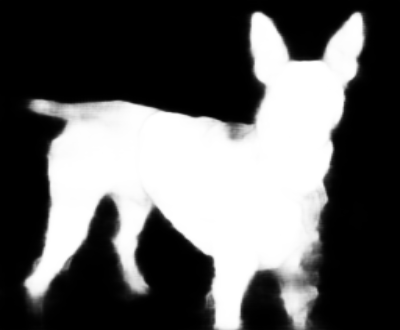}\ &
\includegraphics[width=0.085\linewidth,height=1.25cm]{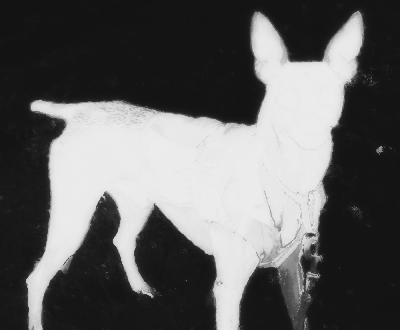}\ &
\includegraphics[width=0.085\linewidth,height=1.25cm]{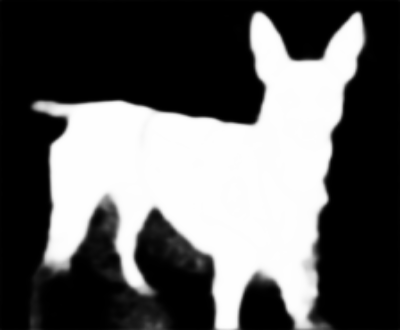}\ &
\includegraphics[width=0.085\linewidth,height=1.25cm]{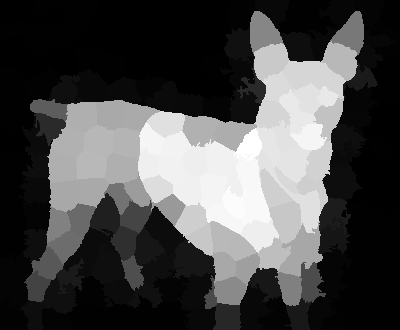}\ &
\includegraphics[width=0.085\linewidth,height=1.25cm]{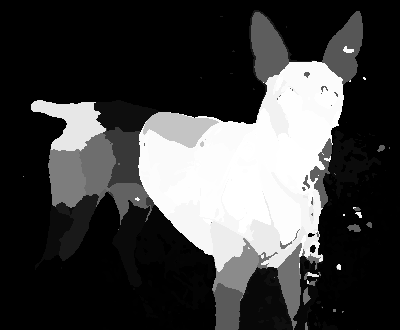}\ &
\includegraphics[width=0.085\linewidth,height=1.25cm]{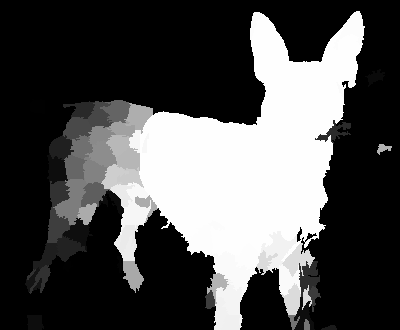}\ &
\includegraphics[width=0.085\linewidth,height=1.25cm]{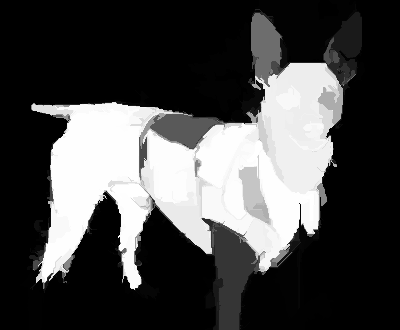}\ &
\includegraphics[width=0.085\linewidth,height=1.25cm]{0070_RFCN.jpg}\ &
\includegraphics[width=0.085\linewidth,height=1.25cm]{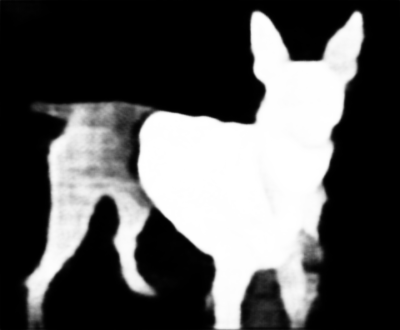}\ \\
\vspace{-0.5mm}
\includegraphics[width=0.085\linewidth,height=1.2cm]{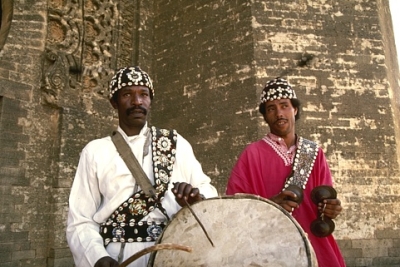}\ &
\includegraphics[width=0.085\linewidth,height=1.2cm]{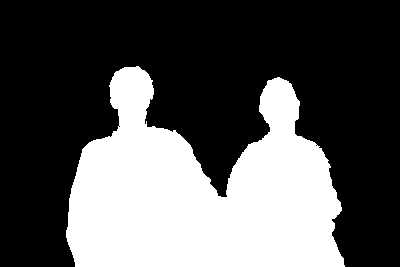}\ &
\includegraphics[width=0.085\linewidth,height=1.2cm]{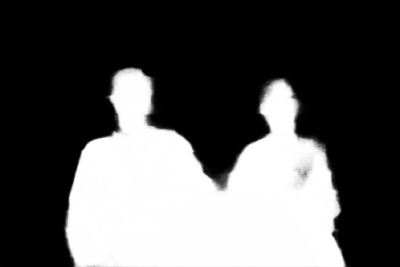}\ &
\includegraphics[width=0.085\linewidth,height=1.2cm]{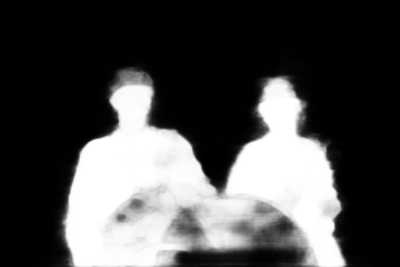}\ &
\includegraphics[width=0.085\linewidth,height=1.2cm]{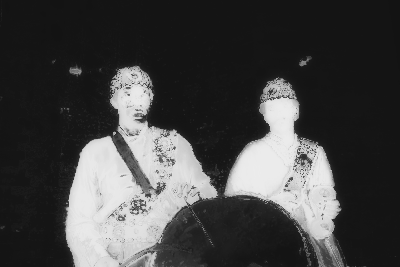}\ &
\includegraphics[width=0.085\linewidth,height=1.2cm]{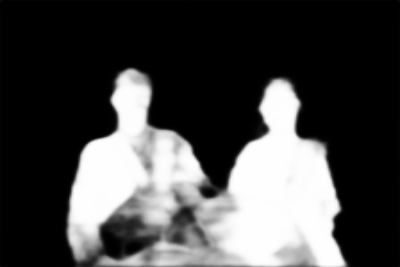}\ &
\includegraphics[width=0.085\linewidth,height=1.2cm]{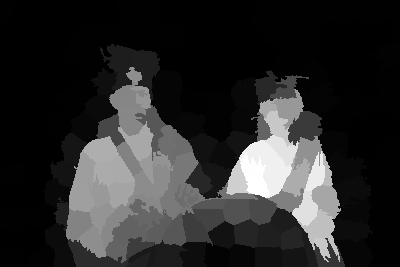}\ &
\includegraphics[width=0.085\linewidth,height=1.2cm]{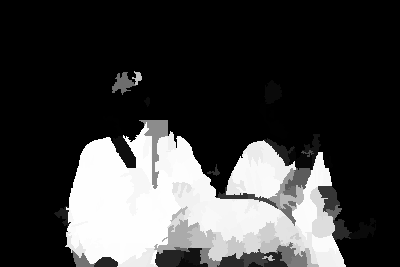}\ &
\includegraphics[width=0.085\linewidth,height=1.2cm]{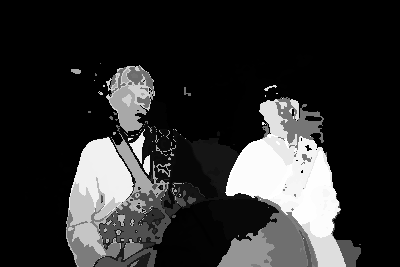}\ &
\includegraphics[width=0.085\linewidth,height=1.2cm]{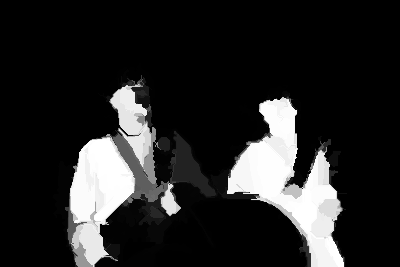}\ &
\includegraphics[width=0.085\linewidth,height=1.2cm]{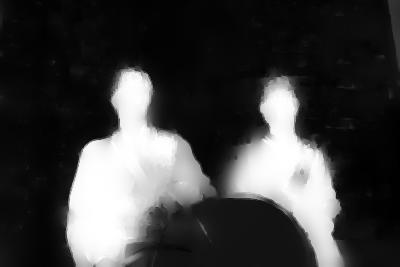}\ &
\includegraphics[width=0.085\linewidth,height=1.2cm]{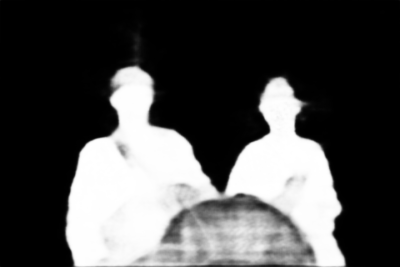}\ \\
\vspace{-0.5mm}
\includegraphics[width=0.085\linewidth,height=1.2cm]{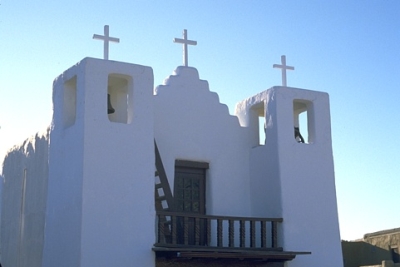}\ &
\includegraphics[width=0.085\linewidth,height=1.2cm]{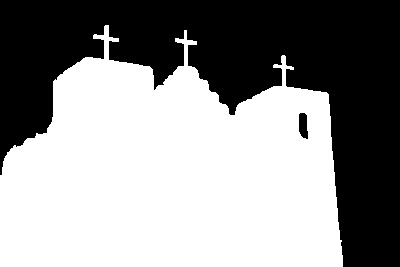}\ &
\includegraphics[width=0.085\linewidth,height=1.2cm]{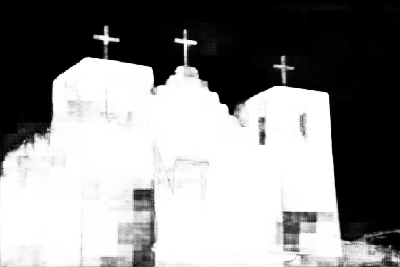}\ &
\includegraphics[width=0.085\linewidth,height=1.2cm]{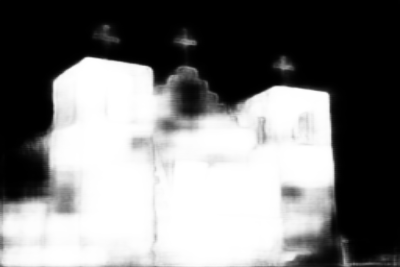}\ &
\includegraphics[width=0.085\linewidth,height=1.2cm]{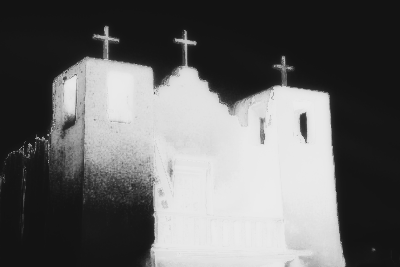}\ &
\includegraphics[width=0.085\linewidth,height=1.2cm]{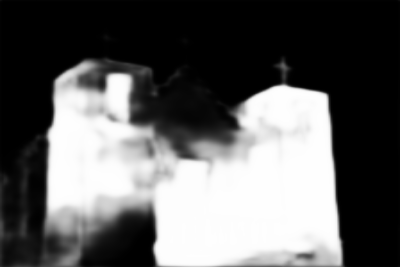}\ &
\includegraphics[width=0.085\linewidth,height=1.2cm]{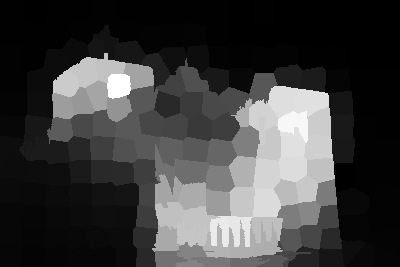}\ &
\includegraphics[width=0.085\linewidth,height=1.2cm]{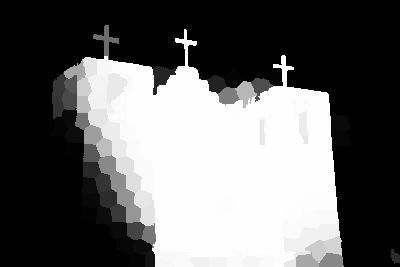}\ &
\includegraphics[width=0.085\linewidth,height=1.2cm]{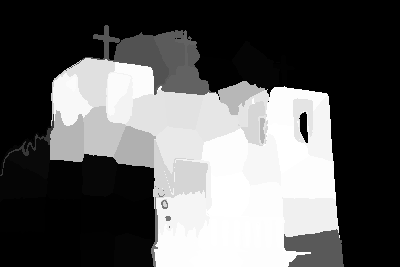}\ &
\includegraphics[width=0.085\linewidth,height=1.2cm]{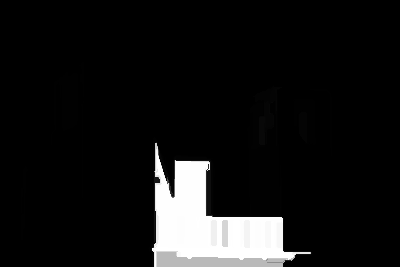}\ &
\includegraphics[width=0.085\linewidth,height=1.2cm]{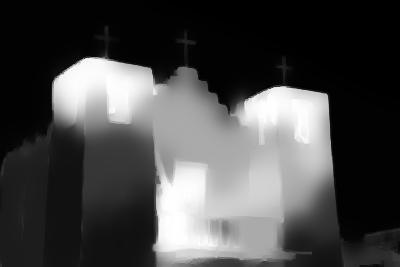}\ &
\includegraphics[width=0.085\linewidth,height=1.2cm]{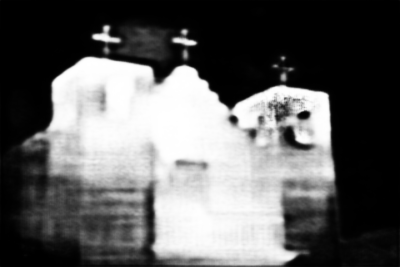}\ \\
\vspace{-0.5mm}
\includegraphics[width=0.085\linewidth,height=1.2cm]{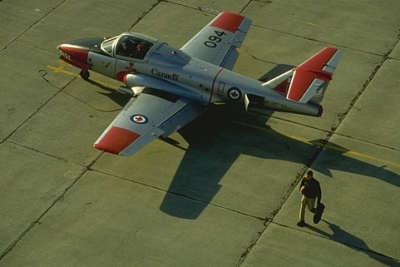}\ &
\includegraphics[width=0.085\linewidth,height=1.2cm]{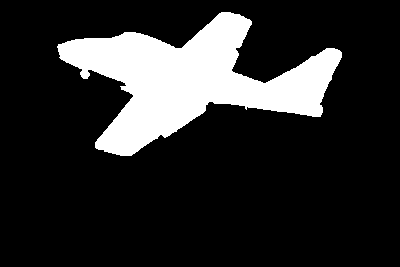}\ &
\includegraphics[width=0.085\linewidth,height=1.2cm]{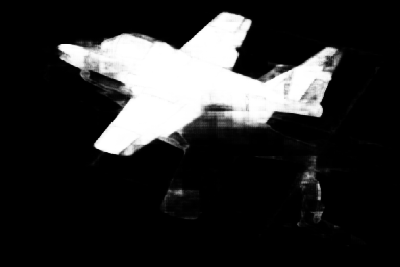}\ &
\includegraphics[width=0.085\linewidth,height=1.2cm]{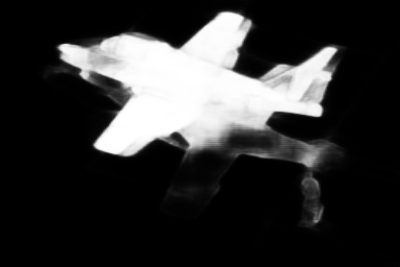}\ &
\includegraphics[width=0.085\linewidth,height=1.2cm]{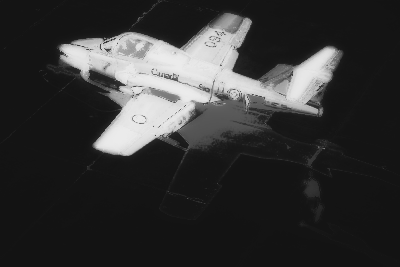}\ &
\includegraphics[width=0.085\linewidth,height=1.2cm]{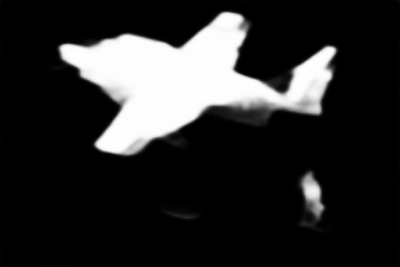}\ &
\includegraphics[width=0.085\linewidth,height=1.2cm]{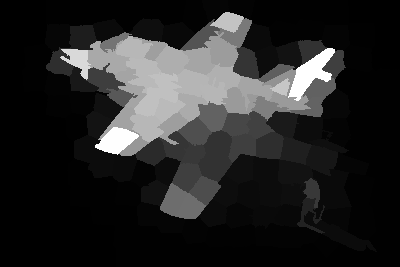}\ &
\includegraphics[width=0.085\linewidth,height=1.2cm]{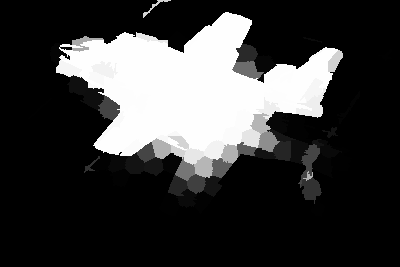}\ &
\includegraphics[width=0.085\linewidth,height=1.2cm]{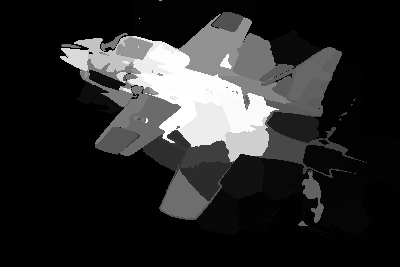}\ &
\includegraphics[width=0.085\linewidth,height=1.2cm]{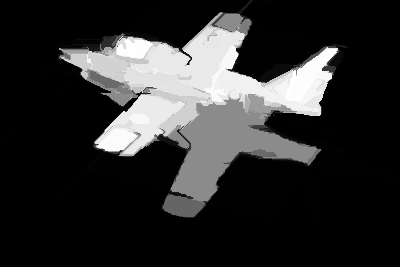}\ &
\includegraphics[width=0.085\linewidth,height=1.2cm]{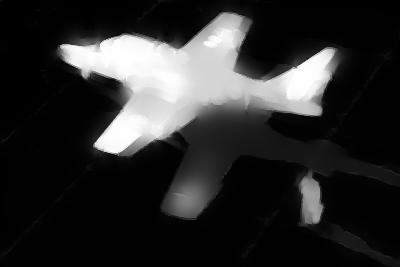}\ &
\includegraphics[width=0.085\linewidth,height=1.2cm]{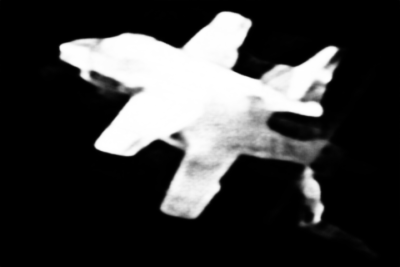}\ \\
\vspace{-0.5mm}
\includegraphics[width=0.085\linewidth,height=1.2cm]{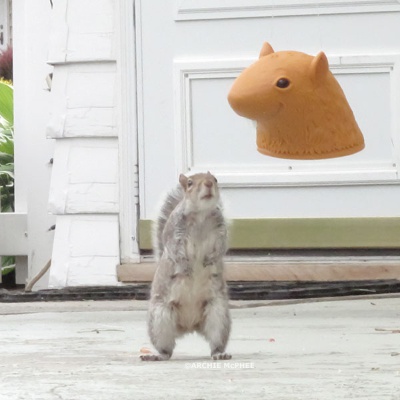}\ &
\includegraphics[width=0.085\linewidth,height=1.2cm]{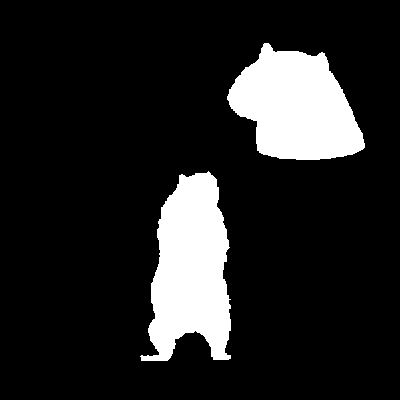}\ &
\includegraphics[width=0.085\linewidth,height=1.2cm]{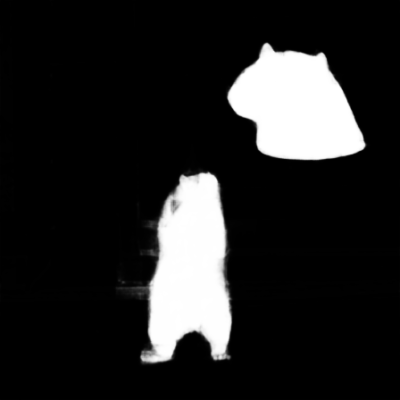}\ &
\includegraphics[width=0.085\linewidth,height=1.2cm]{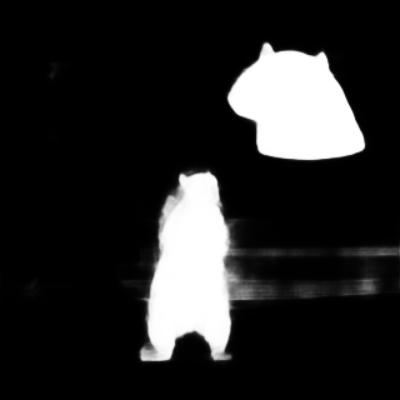}\ &
\includegraphics[width=0.085\linewidth,height=1.2cm]{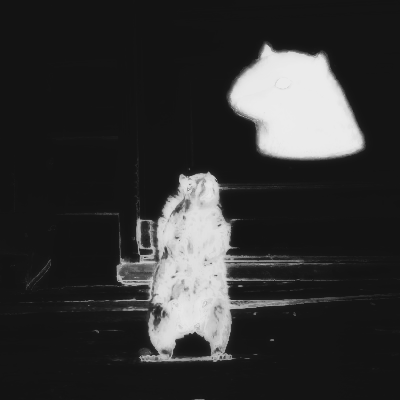}\ &
\includegraphics[width=0.085\linewidth,height=1.2cm]{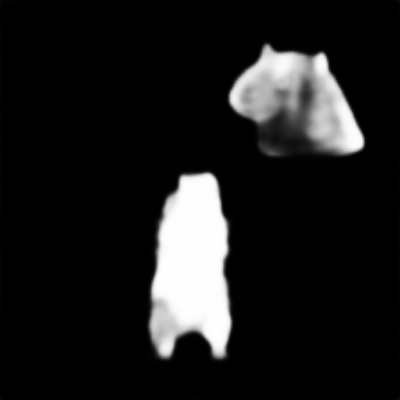}\ &
\includegraphics[width=0.085\linewidth,height=1.2cm]{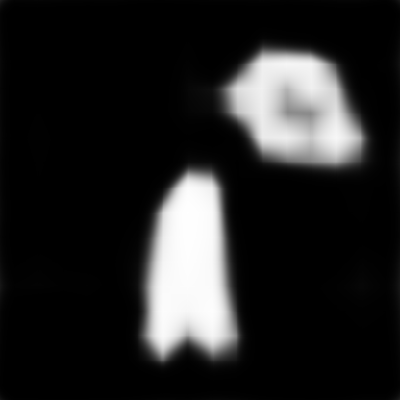}\ &
\includegraphics[width=0.085\linewidth,height=1.2cm]{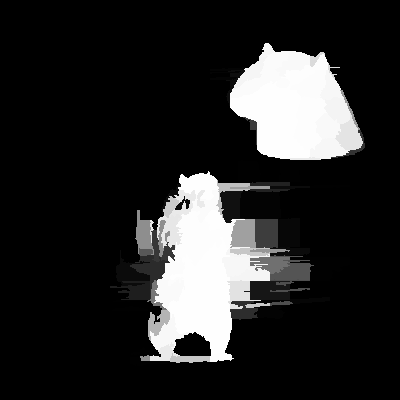}\ &
\includegraphics[width=0.085\linewidth,height=1.2cm]{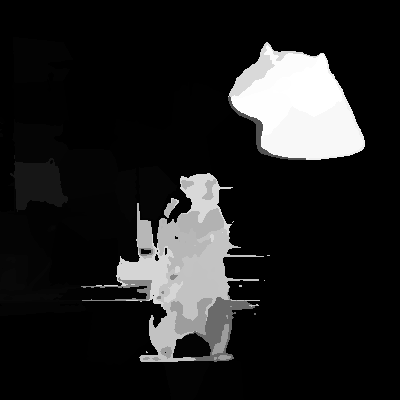}\ &
\includegraphics[width=0.085\linewidth,height=1.2cm]{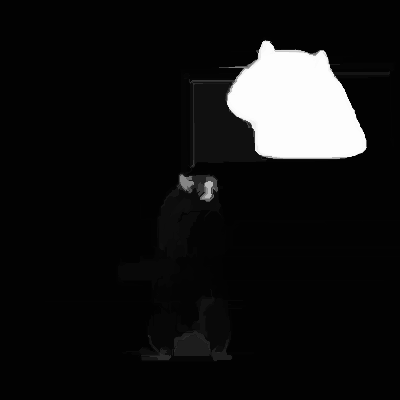}\ &
\includegraphics[width=0.085\linewidth,height=1.2cm]{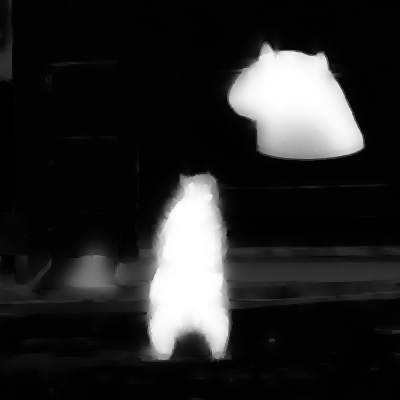}\ &
\includegraphics[width=0.085\linewidth,height=1.2cm]{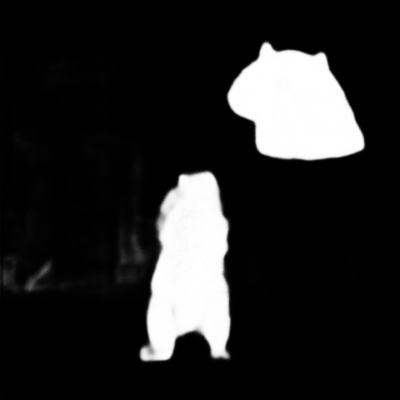}\ \\
{\small (a)} & {\small(b)} & {\small(c)} & {\small(d)} & {\small(e)}& {\small(f)}& {\small(g)}
& {\small(h)}& {\small(i)}& {\small(j)}& {\small(k)}& {\small(l)}\ \\
\end{tabular}
}
\vspace{-2mm}
\caption{Comparison of saliency maps. (a) Input images; (b) Ground truth; (c) Ours; (d) AMU~\cite{zhang2017amulet}; (e) DCL~\cite{li2016dcl}; (f) DHS~\cite{liu2016dhsnet}; (g) DS~\cite{li2016ds}; (h) ELD~\cite{lee2016deep}; (i) MCDL~\cite{zhao2015saliency}; (j) MDF~\cite{li2015visual}; (k) RFCN~\cite{wang2016saliency}; (l) UCF~\cite{zhang2017learning}. The results of LEGS~\cite{wang2015deep}, BL~\cite{tong2015salient}, BSCA~\cite{qin2015saliency}, DRFI~\cite{jiang2013salient} and DSR~\cite{li2013saliency} can be found in the supplemental material.
\label{fig:map_comparison}}
\vspace{-6mm}
\end{figure*}
\begin{figure*}
\centering
\resizebox{1\textwidth}{!}
{
\begin{tabular}{@{}c@{}c@{}c@{}c@{}c@{}c@{}c@{}c@{}c@{}c@{}c@{}c}
\vspace{-0.5mm}
\includegraphics[width=0.085\linewidth,height=1.25cm]{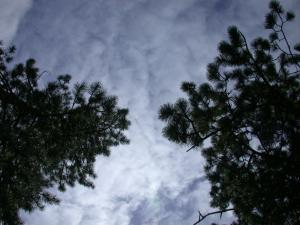}\ &
\includegraphics[width=0.085\linewidth,height=1.25cm]{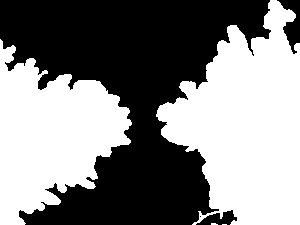}\ &
\includegraphics[width=0.085\linewidth,height=1.25cm]{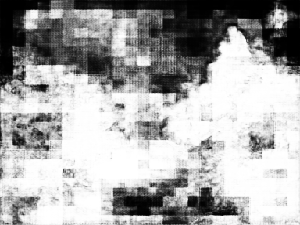}\ &
\includegraphics[width=0.085\linewidth,height=1.25cm]{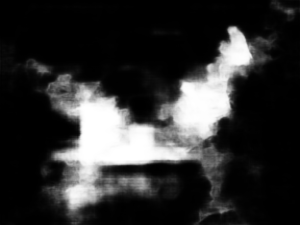}\ &
\includegraphics[width=0.085\linewidth,height=1.25cm]{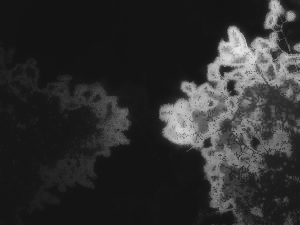}\ &
\includegraphics[width=0.085\linewidth,height=1.25cm]{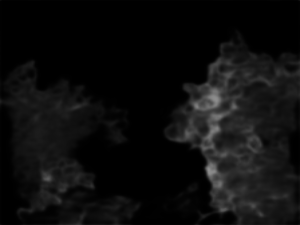}\ &
\includegraphics[width=0.085\linewidth,height=1.25cm]{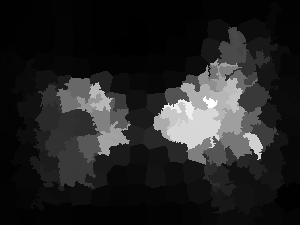}\ &
\includegraphics[width=0.085\linewidth,height=1.25cm]{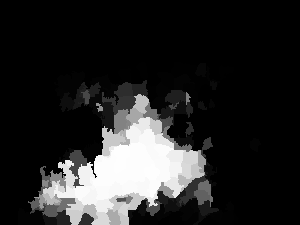}\ &
\includegraphics[width=0.085\linewidth,height=1.25cm]{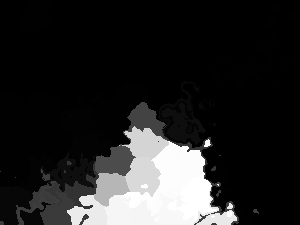}\ &
\includegraphics[width=0.085\linewidth,height=1.25cm]{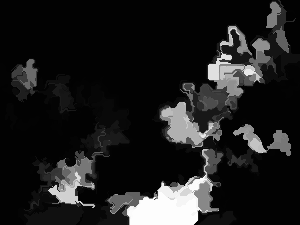}\ &
\includegraphics[width=0.085\linewidth,height=1.25cm]{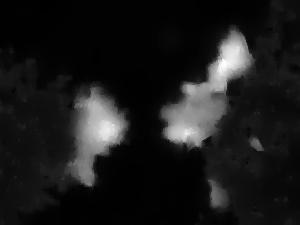}\ &
\includegraphics[width=0.085\linewidth,height=1.25cm]{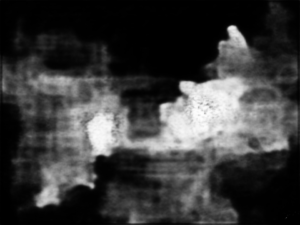}\ \\
\vspace{-0.5mm}
\includegraphics[width=0.085\linewidth,height=1.25cm]{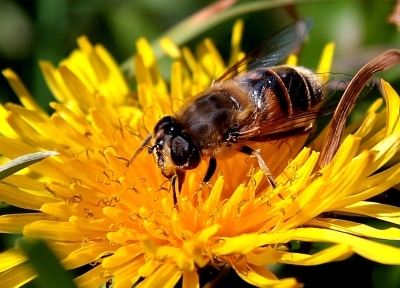}\ &
\includegraphics[width=0.085\linewidth,height=1.25cm]{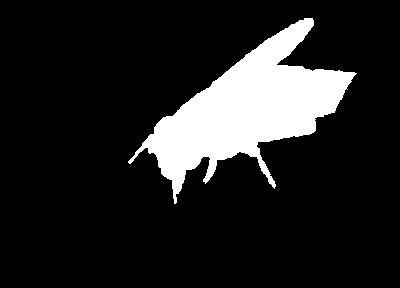}\ &
\includegraphics[width=0.085\linewidth,height=1.25cm]{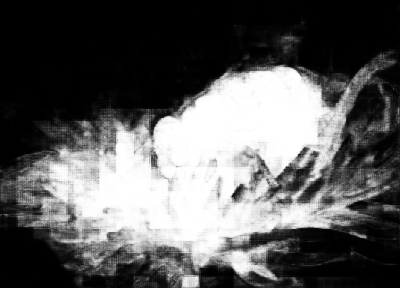}\ &
\includegraphics[width=0.085\linewidth,height=1.25cm]{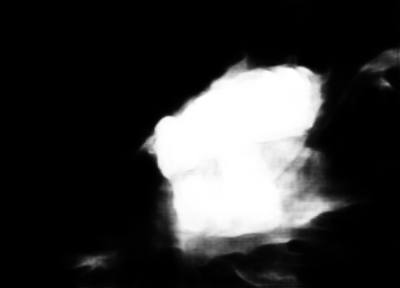}\ &
\includegraphics[width=0.085\linewidth,height=1.25cm]{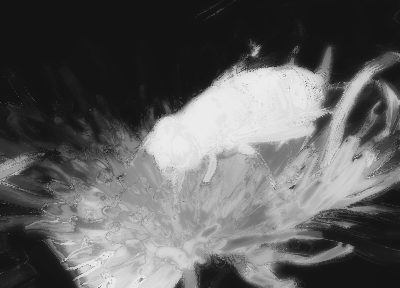}\ &
\includegraphics[width=0.085\linewidth,height=1.25cm]{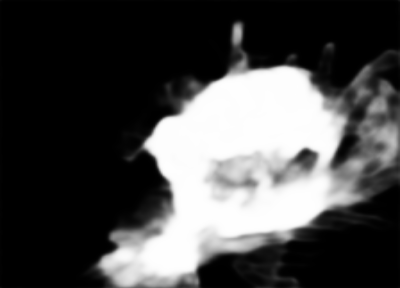}\ &
\includegraphics[width=0.085\linewidth,height=1.25cm]{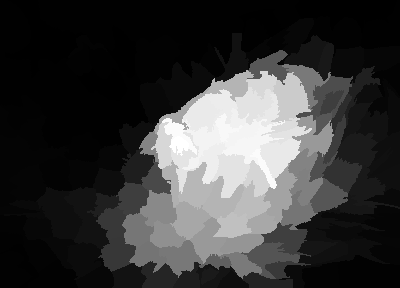}\ &
\includegraphics[width=0.085\linewidth,height=1.25cm]{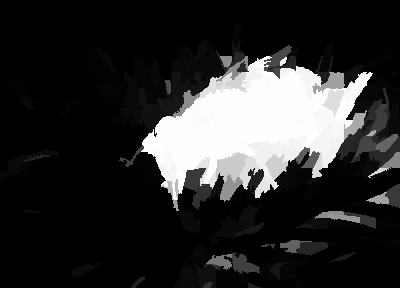}\ &
\includegraphics[width=0.085\linewidth,height=1.25cm]{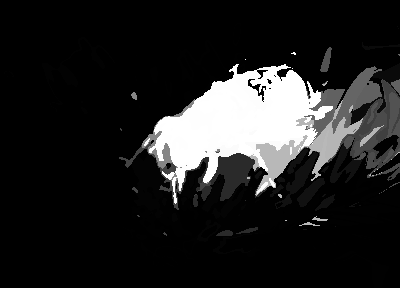}\ &
\includegraphics[width=0.085\linewidth,height=1.25cm]{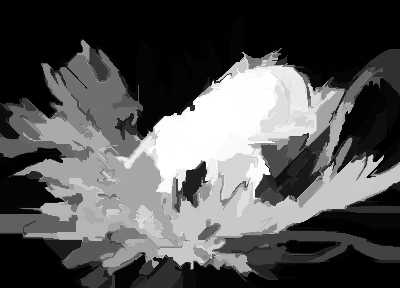}\ &
\includegraphics[width=0.085\linewidth,height=1.25cm]{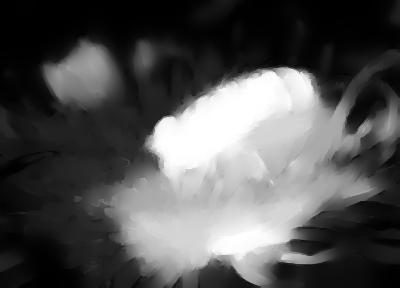}\ &
\includegraphics[width=0.085\linewidth,height=1.25cm]{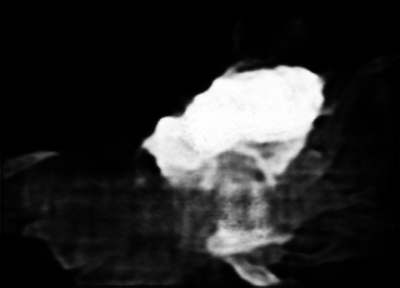}\ \\
\vspace{-0.5mm}
\includegraphics[width=0.085\linewidth,height=1.25cm]{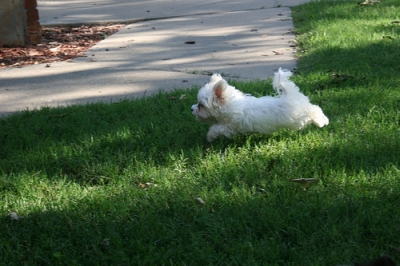}\ &
\includegraphics[width=0.085\linewidth,height=1.25cm]{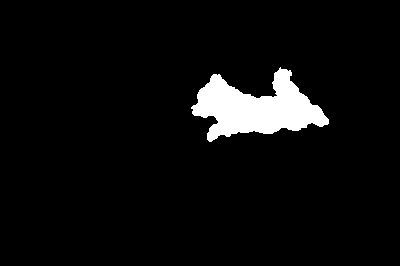}\ &
\includegraphics[width=0.085\linewidth,height=1.25cm]{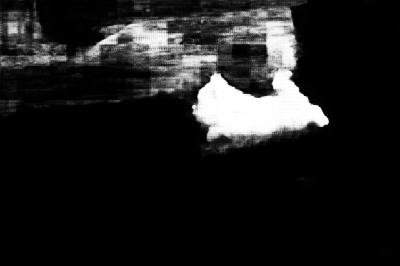}\ &
\includegraphics[width=0.085\linewidth,height=1.25cm]{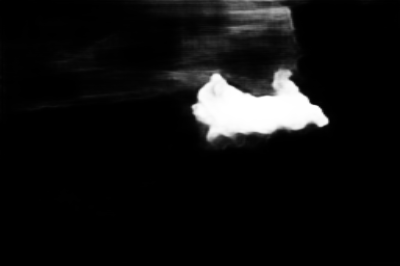}\ &
\includegraphics[width=0.085\linewidth,height=1.25cm]{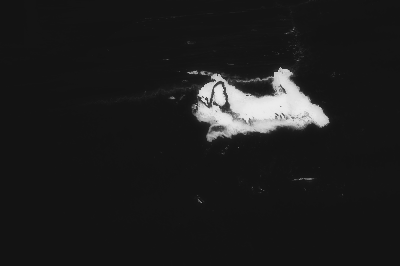}\ &
\includegraphics[width=0.085\linewidth,height=1.25cm]{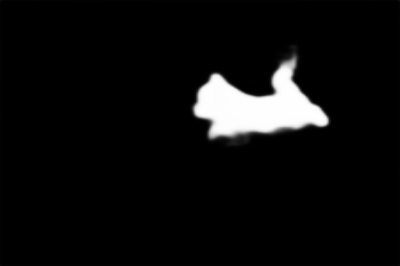}\ &
\includegraphics[width=0.085\linewidth,height=1.25cm]{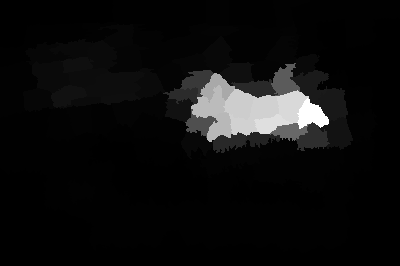}\ &
\includegraphics[width=0.085\linewidth,height=1.25cm]{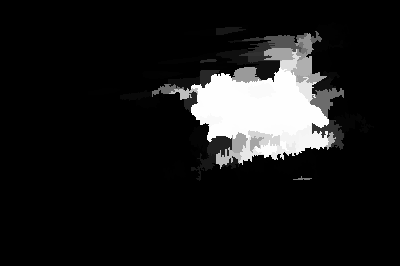}\ &
\includegraphics[width=0.085\linewidth,height=1.25cm]{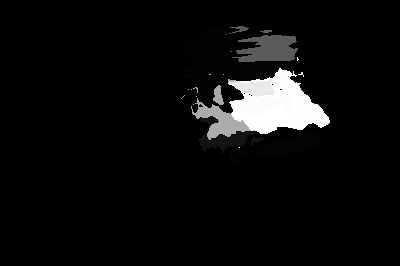}\ &
\includegraphics[width=0.085\linewidth,height=1.25cm]{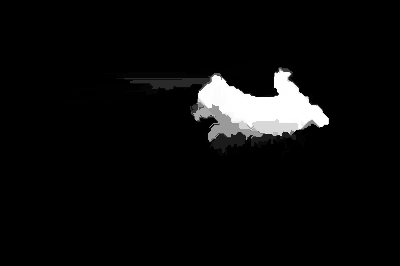}\ &
\includegraphics[width=0.085\linewidth,height=1.25cm]{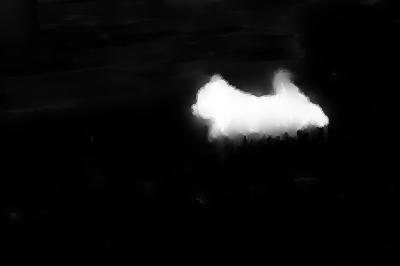}\ &
\includegraphics[width=0.085\linewidth,height=1.25cm]{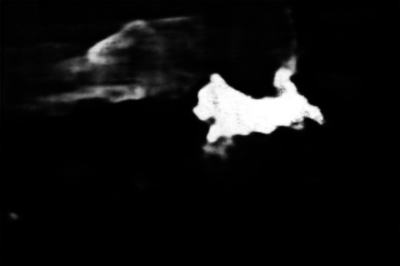}\ \\
{\small (a)} & {\small(b)} & {\small(c)} & {\small(d)} & {\small(e)}& {\small(f)}& {\small(g)}
& {\small(h)}& {\small(i)}& {\small(j)}& {\small(k)}& {\small(l)}\ \\
\end{tabular}
}
\vspace{-2mm}
\caption{Failure examples. (a) Input images; (b) Ground truth; (c) Ours. Top-ranked results: (d) AMU~\cite{zhang2017amulet}; (e) DCL~\cite{li2016dcl}; (f) DHS~\cite{liu2016dhsnet}; (g) DS~\cite{li2016ds}; (h) ELD~\cite{lee2016deep}; (i) MCDL~\cite{zhao2015saliency}; (j) MDF~\cite{li2015visual}; (k) RFCN~\cite{wang2016saliency}; (l) UCF~\cite{zhang2017learning}.
\label{fig:failure_comparison}}
\vspace{-6mm}
\end{figure*}
\vspace{-6mm}
{\flushleft\textbf{Quantitative Results.}}
Fig.~\ref{fig:map_comparison} provides several visual examples for qualitative comparisons.
In various challenging conditions, our method consistently outperforms other compared methods.
For example, salient objects contain inconsistent regions (the 1th-3th row), salient objects touches the image boundaries (the 1th-4th row), the background is complex and confusing (the 1th-2th, 4th-5th row) and multiple salient objects (the 4th, 7th row).
%
%
Our model can accurately locate the salient objects and simultaneously capture clear object boundaries, generating coherent and precise saliency maps effectively.
In addition, we observe that our model can highlight the salient objects under multi-contrast and shadow cases (the 1-2th, 6th row).
By contrast, other models tend to be ineffective on these challenging conditions due to the lack of image reflection and multi-source fusion strategies.
%
%
Fig.~\ref{fig:failure_comparison} shows some failure examples.
When the salient objects have scattered details (the 1th row), our method may detect the bulks as the salient object, but human can easily locate the real objects.
When the salient objects have varied saliency (the 2th row), our method and other approaches simultaneously fail to detect the objects.
In addition, our method may be effected by cluttered light (the 3th row).
\vspace{-4mm}
\subsection{Ablation Studies}
\begin{table*}
\begin{center}
\resizebox{1\textwidth}{!}
{
\begin{tabular}{|c|c|c|c|c|c|c|c|}
\hline
Models &\small{Input Fusion} &\small{Early Fusion}&\small{Late Fusion}&\small{Ad-hoc Fusion}&\small{HyperFusion-Net}+RGB &\small{HyperFusion-Net}+TR \\
\hline
$F_\eta$       &0.804&0.821& 0.855& \textcolor[rgb]{0,1,0}{0.863}& \textcolor[rgb]{0,0,1}{0.852}&\textcolor[rgb]{1,0,0}{0.886}        \\
\hline
$MAE$          &0.140&0.129& 0.121& \textcolor[rgb]{0,0,1}{0.074}& \textcolor[rgb]{0,1,0}{0.069}&\textcolor[rgb]{1,0,0}{0.050}       \\
\hline
$S_\lambda$    &0.794&0.814& 0.849& \textcolor[rgb]{0,0,1}{0.860}& \textcolor[rgb]{0,1,0}{0.872}&\textcolor[rgb]{1,0,0}{0.903}       \\
\hline
\end{tabular}
}
\end{center}
\vspace{-2mm}
\caption{Results with different fusion methods w/wo image reflection on the ECSSD dataset. The best three results are shown in \textcolor[rgb]{1,0,0}{red},~\textcolor[rgb]{0,1,0}{green} and \textcolor[rgb]{0,0,1}{blue}, respectively.}
\label{table:aggregation}
\vspace{-6mm}
\end{table*}
With different model settings, we also evaluate the performance of main components in our model.
All models are trained on the augmented MSRA10K dataset and share the same hyper-parameters described in subsection 4.2.
Due to the limitation of space, we only show the results on the ECSSD dataset.
Other datasets have the similar performance trend.
%
\vspace{-2mm}
{\flushleft\textbf{The effect of different losses.}}
Tab.~\ref{table:loss} shows the experimental results with different losses.
From the results, we can see that the ICNN only using the channel concatenation operator without H-Fusion (model (a)) has achieved comparable performance to most deep learning methods.
This confirms the effectiveness of reflection features.
With the H-Fusion, the resulting ICNN (model (b)) improves the performance by a large margin.
The main reason is that the fusion method introduces more complementary information, which helps to locate the salient objects.
In addition, it is no wonder that training with the $\mathcal{L}_{wbce}$ loss achieves better results than $\mathcal{L}_{bce}$.
With the structure perceptual loss $\mathcal{L}_{sp}$, the model achieves better performance in terms of S-measure.
When taking them together, the model achieves best results in all evaluation metrics.
\vspace{-2mm}
{\flushleft\textbf{The effect of fusion methods.}}
To verify the benefits of fusion methods, we also compare our fusion strategy with the methods described in Subsection 2.2.
For the input fusion, we concatenate the image pair in channel-wise and use the SegNet~\cite{badrinarayanan2015segnet2} for SOD.
For other fusion methods, we follow the practice of previous works~\cite{wang2015deep,zhao2015saliency,li2015visual,li2016dcl,zhang2017amulet}, and use the VGG-16 model to build the fusion models.
More details are listed in the supplementary material.
Tab.~\ref{table:aggregation} shows the quantitative results.
From the results, we can see that adding fusing points can consistently improve the SOD performance.
This fact confirms our motivations and claims.
\vspace{-2mm}
{\flushleft\textbf{The effect of image separation.}}
Tab.~\ref{table:aggregation} also provides the results of the models with/without image reflection separation.
The results indicates the benefits of image reflection separation across all the evaluation metrics.
Besides, image reflection separation shows significant improvements in S-measure.
The main reason is that our reflection separation is capable of: 1) retaining the main content and structure of RGB images and  disregarding local salient distractions; 2) highlighting local details of the target salient object with the reflected views.
%
\vspace{-4mm}
\section{Conclusion and Future Work}
\vspace{-2mm}
In this work, we introduce a novel end-to-end feature learning framework for SOD.
Our method utilizes a densely hierarchical feature fusion network, named \emph{HyperFusion-Net}, to predict the most important area and segment the associated objects.
Inspired by the human perception system, we first decompose input images into reflective image pairs by content-preserving transforms.
Then, the complementary features of reflective image pairs can be jointly extracted by an interweaved CNN and hierarchically combined with a hyper-dense fusion mechanism.
Based on the fused multi-scale features, our method finally achieves a promising way of predicting SOD.
Extensive experiments demonstrate that the proposed method achieves significant improvement over the baseline with a large margin, and performs better than other state-of-the-art methods.

Based on the superior performance and flexibility, we plan to apply the framework to other multi-modal applications, for example, RGB-D SOD.
%
It is also promising to leverage this framework to fuse other modalities, such as image and text for image capturing, image and audio for video classification and so on.
\clearpage

\bibliographystyle{splncs}
\bibliography{egbib}
\end{document}